\def\paperID{1378}
\def\confName{CVPR}
\def\confYear{2025}

\def\authorBlock{
    Yunze Liu\textsuperscript{1,2} \qquad 
    Li Yi\textsuperscript{1,3,2}  \qquad
    \\
    \textsuperscript{1}IIIS, Tsinghua University \quad
    \textsuperscript{2}Shanghai Qi Zhi Institute 
    \quad
    \textsuperscript{3}Shanghai Artificial Intelligence Laboratory 
}

\newif\ifreview 
\newif\ifarxiv \newcommand{\arxiv}{\arxivtrue}
\newif\ifcamera 
\newif\ifrebuttal 

\arxiv

\pdfoutput=1
\documentclass[10pt,twocolumn,letterpaper]{article}
\ifreview \usepackage[review]{cvpr} \fi
\ifarxiv \usepackage[pagenumbers]{cvpr} \fi
\ifrebuttal \usepackage[rebuttal]{cvpr} \fi
\ifcamera \usepackage{cvpr} \fi

\newcommand{\rblue}[1]{\textcolor{blue}{#1}}
\usepackage{graphicx}	
\usepackage{amsmath}	
\usepackage{amssymb}	
\usepackage{booktabs}
\usepackage{times}
\usepackage{microtype}
\usepackage{epsfig}
\usepackage{caption}
\usepackage{float}
\usepackage{placeins}
\usepackage{color, colortbl}
\usepackage{xcolor}
\usepackage{stfloats}
\usepackage{enumitem}
\usepackage{tabularx}
\usepackage{xstring}
\usepackage{multirow}
\usepackage{xspace}
\usepackage{url}
\usepackage{subcaption}

\usepackage[hang,flushmargin]{footmisc}
\usepackage{graphicx}
\usepackage{amsmath}
\usepackage{amssymb}
\usepackage{booktabs}
\usepackage{algorithmic} %
\usepackage{color,xcolor}
\usepackage{amsmath}
\usepackage{xspace}
\usepackage{multirow}
\definecolor{shapecolor}{rgb}{0.0,0.5,0.0}
\definecolor{mygray}{gray}{.9}

\usepackage{hyperref}       %
\usepackage{url}            %
\usepackage{booktabs}       %
\usepackage{graphicx}
\usepackage{amsfonts}       %
\usepackage{nicefrac}       %
\usepackage{microtype}      %
\usepackage{xcolor}         %
\usepackage{algorithm}
\usepackage{tabularx}

\newcommand{\app}{\raise.17ex\hbox{$\scriptstyle\sim$}}

\newcommand{\bs}{\textbf}
\newcommand{\bv}{\textbf}
\newcommand{\br}[1]{\textbf{#1}}

\usepackage{booktabs}   
\usepackage{multirow}   
\usepackage{xcolor}

\usepackage[capitalize]{cleveref}
\crefname{section}{Sec.}{Secs.}
\Crefname{section}{Section}{Sections}
\Crefname{table}{Table}{Tables}
\crefname{table}{Tab.}{Tabs.}
\ifcamera \usepackage[accsupp]{axessibility} \fi

% ...

\ifarxiv  \fi

\newcommand{\R}[1]{{%
    \textbf{%
        \ifstrequal{#1}{1}{\textcolor{red}{R#1}}{%
        \ifstrequal{#1}{2}{\textcolor{blue}{R#1}}{%
        \ifstrequal{#1}{3}{\textcolor{magenta}{R#1}}{%
        \ifstrequal{#1}{4}{\textcolor{teal}{R#1}}{%
                           \textcolor{cyan}{R#1}%
        }}}}%
    }%
}}

\usepackage{xr-hyper}

\makeatletter
\newcommand*{\addFileDependency}[1]{
  \typeout{(#1)}
  \@addtofilelist{#1}
  \IfFileExists{#1}{}{\typeout{No file #1.}}
}

\makeatother

\definecolor{cvprblue}{rgb}{0.21,0.49,0.74}

\usepackage[accsupp]{axessibility}  

\begin{document}
\title{MAP: Unleashing Hybrid Mamba-Transformer Vision Backbone's Potential with Masked Autoregressive Pretraining}
\author{\authorBlock}
\maketitle
\begin{abstract}
Hybrid Mamba-Transformer networks have recently garnered broad attention. These networks can leverage the scalability of Transformers while capitalizing on Mamba's strengths in long-context modeling and computational efficiency. However, the challenge of effectively pretraining such hybrid networks remains an open question. Existing methods, such as Masked Autoencoders (MAE) or autoregressive (AR) pretraining, primarily focus on single-type network architectures. In contrast, pretraining strategies for hybrid architectures must be effective for both Mamba and Transformer components. Based on this, we propose Masked Autoregressive Pretraining (MAP) to pretrain a hybrid Mamba-Transformer vision backbone network. This strategy combines the strengths of both MAE and Autoregressive pretraining, improving the performance of Mamba and Transformer modules within a unified paradigm. Experimental results show that the hybrid Mamba-Transformer vision backbone network pretrained with MAP significantly outperforms other pretraining strategies, achieving state-of-the-art performance. We validate the method's effectiveness on both 2D and 3D datasets and provide detailed ablation studies to support the design choices for each component. Code and checkpoints are available at \href{https://github.com/yunzeliu/MAP}{https://github.com/yunzeliu/MAP}

\end{abstract}

\section{Introduction}
\label{sec:intro}

The Mamba-Transformer backbone\cite{hatamizadeh2024mambavision, lieber2024jamba, wang2024longllava, chen2024maskmamba} has recently attracted widespread attention.  It leverages the scalability advantages of Transformers and utilizes Mamba's\citep{gu2023mamba,zhuvision} strong capabilities in long-context language modeling. However, to scale up Mamba-Transformer vision backbones, an effective pretraining strategy is essential for maximizing the combined capabilities of Mamba and Transformer. Our work aims to take the first step in this direction.

\begin{figure}[h]
    \centering
    \includegraphics[width=0.8\linewidth]{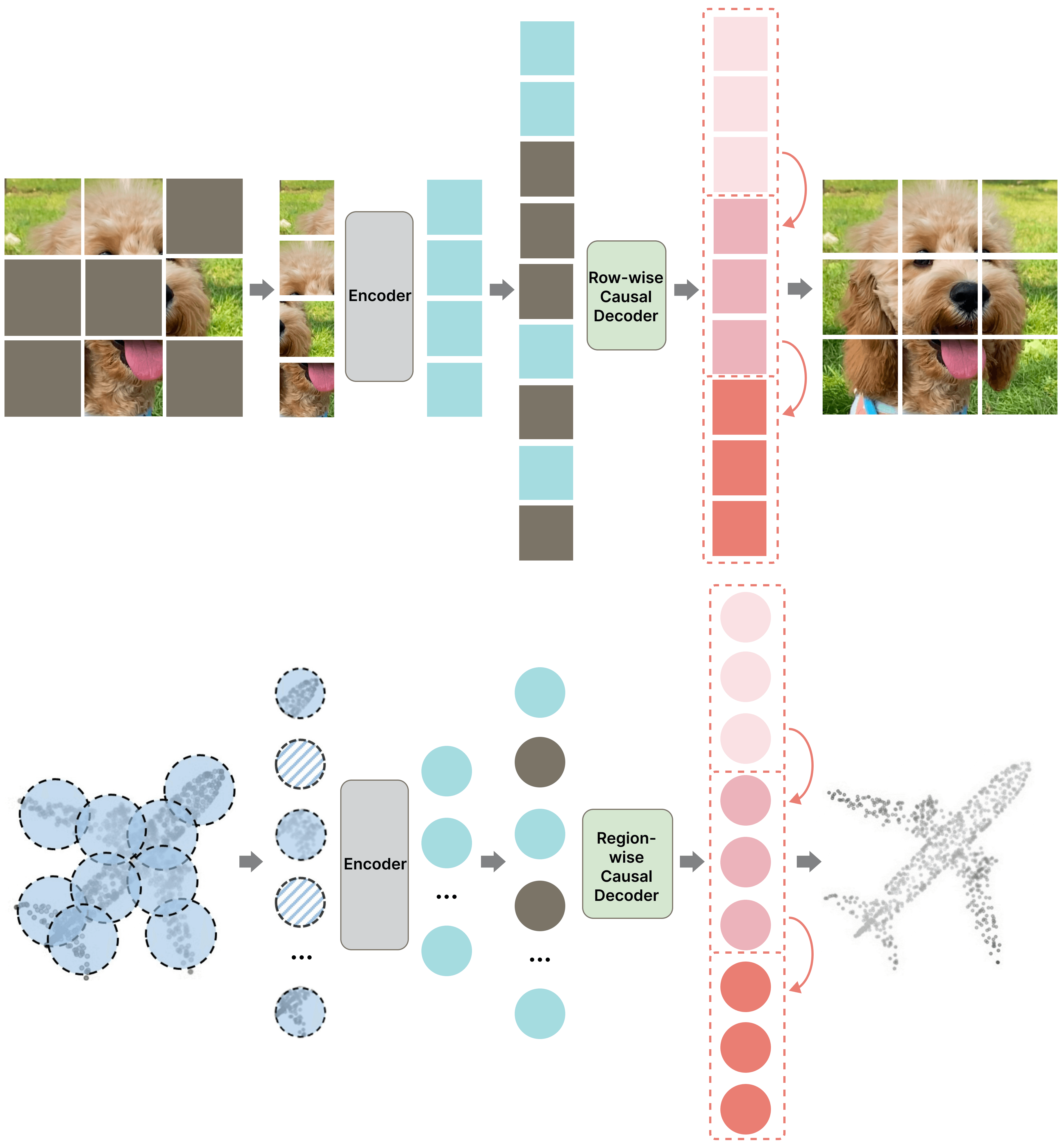}
    \vspace{-3mm}
    \caption{We propose \textbf{M}asked \textbf{A}utoregressive \textbf{P}retraining to pretrain the hybrid Mamba-Transformer backbone. It demonstrates significant performance improvements on both 2D and 3D tasks.}
    \label{fig:teaser_head}
    \vspace{-5mm}
\end{figure}

\begin{figure*}[h]
    \centering
    \includegraphics[width=0.8\linewidth]{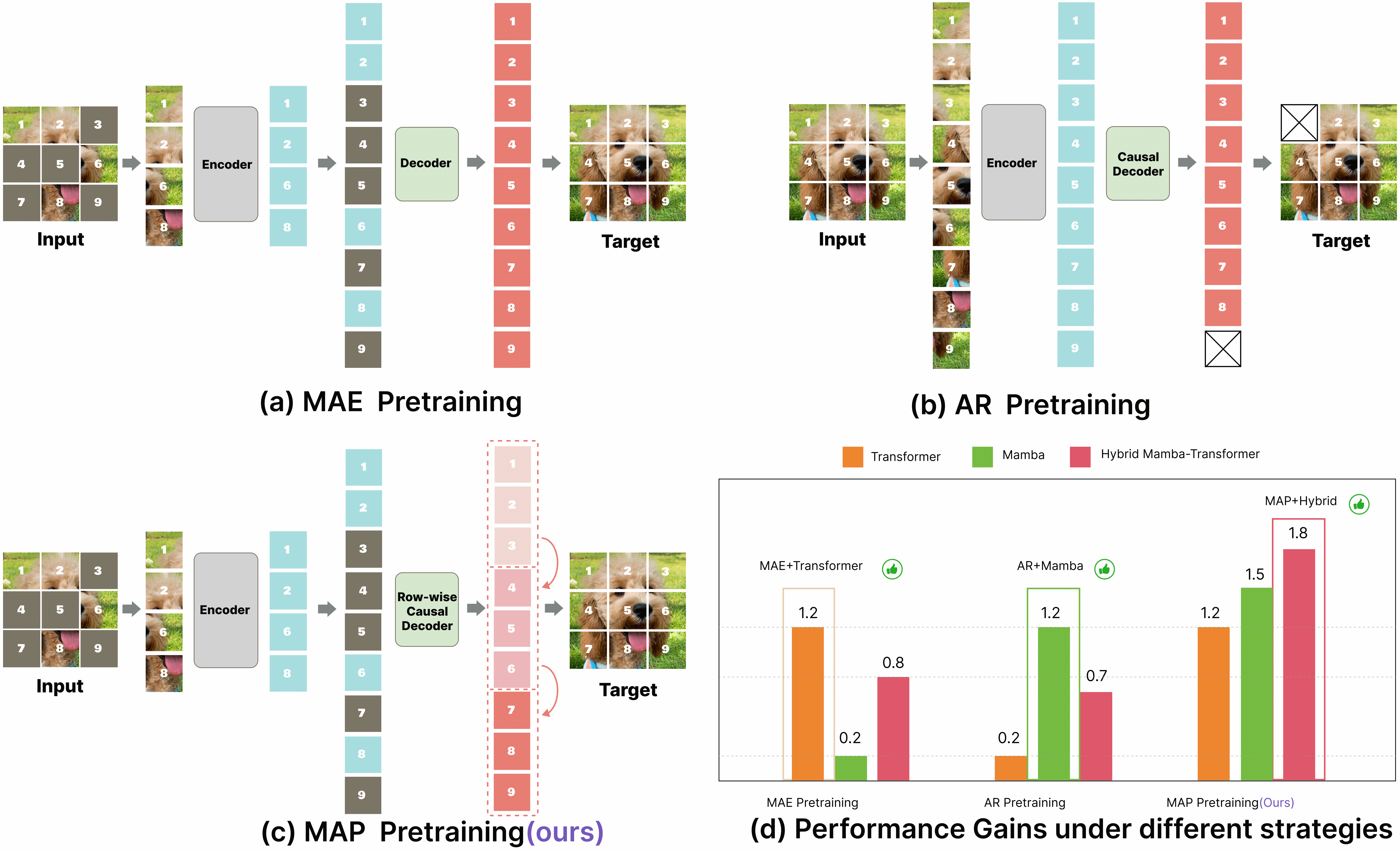}
    \vspace{-3mm}
    \caption{\textbf{(a) MAE Pretraining.} Its core lies in reconstructing the masked tokens based on the unmasked tokens to build a global bidirectional contextual understanding. 
    \textbf{(b) AR Pretraining.} It focuses on building correlations between contexts, and its scalability has been thoroughly validated in the field of large language models.
    \textbf{(c) MAP Pretraining(ours).} Our method first randomly masks the input image, and then reconstructs the original image in a row-by-row autoregressive manner. This pretraining approach demonstrates significant advantages in modeling contextual features of local characteristics and the correlations between local features, making it highly compatible with the Mamba-Transformer hybrid architecture. 
    \textbf{(d) Performance Gains under different pretraining strategies on ImageNet-1K.} We found MAE pretraining is better suited for Transformers, while AR is more compatible with Mamba. MAP, on the other hand, is more suited for the Mamba-Transformer backbone. Additionally, MAP also demonstrates impressive performance when pretraining with pure Mamba or pure Transformer backbones, showcasing the effectiveness and broad applicability of our method.}
    \label{fig:teaser}
    \vspace{-5mm}
\end{figure*}

Developing an effective pretraining strategy for Mamba-Transformer vision backbones is challenging. Although the capability of MAE in pretraining Transformers has been well validated, pretraining methods for Mamba are still underexplored, and the optimal approach remains unclear. Additionally, the hybrid structure requires a pretraining strategy compatible with both computation blocks. This is particularly challenging because the State Space Model captures visual features very differently from Transformers.

To address these challenges, we conducted pilot studies and identified three key observations.
Firstly, existing popular pretraining strategies for Transformers, such as MAE\citep{he2022masked} and Contrastive Learning\citep{he2020momentum}, do not yield satisfactory results for Mambas.
Secondly, Autoregressive Pretraining\citep{ren2024autoregressive} can be effective for Mamba-based vision backbones with an appropriate scanning pattern and token masking ratio.
Thirdly, pretraining strategies suitable for either Mamba or Transformers do not effectively benefit the other.

Based on the above observations, we develop a novel pretraining strategy suitable for the Mamba-Transformer vision backbone named Masked Autoregressive pretraining, or MAP for short. The key is a hierarchical pretraining objective where local MAE is leveraged to learn good local attention for the Transformer blocks while global autoregressive pretraining enables the Mamba blocks to learn meaningful contextual information.
Specifically, the pretraining method is supported by two key designs. First, we leverage local MAE to enable the hybrid framework, particularly the Transformer module, to learn local bidirectional connectivity. This requires the hybrid network to predict all tokens within a local region after perceiving local bidirectional information.
Second, we autoregressively generate tokens for each local region to allow the hybrid framework, especially the Mamba module, to learn rich contextual information. This requires the network to autoregressively generate subsequent local regions based on the previously decoded tokens. 

Notably, since our focus in this paper is on pretraining strategies for hybrid frameworks rather than the hybrid frameworks themselves, we empirically choose a densely mixed hybrid scheme as the default hybrid framework and refer to it as HybridNet. Our experiments demonstrate that HybridNet pretrained with MAP outperforms other pretraining strategies by a significant margin on the 2D and 3D vision tasks. Extensive ablation studies confirm the rationale and effectiveness of each design choice.

\noindent Our contributions are threefold as follows: \\
\textbf{Firstly,}  we propose a novel method for pretraining the Hybrid Mamba-Transformer Vision Backbone for the first time, enhancing the performance of hybrid backbones as well as pure Mamba and pure Transformer backbones within a unified paradigm. \textbf{Secondly,} For diagnostic purposes, we also conduct an in-depth analysis of the key components of pure-Mamba with autoregressive pretraining, revealing that the effectiveness hinges on maintaining consistency between the pretraining order and the Mamba scanning order, along with an appropriate token masking ratio. \textbf{Thirdly,} we demonstrate that our proposed method, MAP, significantly improves the performance of both Mamba-Transformer and pure Mamba backbones across various 2D and 3D datasets. Extensive ablation studies confirm the rationale and effectiveness of each design choice.

\section{Related Work}
\label{sec:related}
\vspace{-2mm}
\textbf{Vision Mambas and Vision Transformers.}
Vision Mamba(Vim)\citep{vim} is an efficient model for visual representation learning, leveraging bidirectional state space blocks to outperform traditional vision transformers like DeiT in both performance and computational efficiency. The VMamba\citep{vmamba} architecture, built using Visual State-Space blocks and 2D Selective Scanning, excels in visual perception tasks by balancing efficiency and accuracy. Autoregressive pretraining(ARM)\citep{ren2024autoregressive} further boosts Vision Mamba's performance, enabling it to achieve superior accuracy and faster training compared to conventional supervised models. Nevertheless, why autoregression is effective for Vision Mamba and what the key factors are remains an unresolved question. In this paper, we explore the critical design elements behind the success of Mamba's autoregressive pretraining for the first time.
The key difference between our proposed MAP and ARM\citep{ren2024autoregressive} lies in our use of random masking in the AR process, where the network decodes the next complete local information based on the masked local information. The masking mechanism enhances the modeling capability of local information and the relevance of context. Vision Transformers(ViT)\citep{dosovitskiy2020image} adapt transformer architectures to image classification by treating image patches as sequential tokens. Swin Transformer\citep{liu2021swin} introduces a hierarchical design with shifted windows, effectively capturing both local and global information for image recognition. MAE \citep{he2022masked} enhances vision transformers through self-supervised learning, where the model reconstructs masked image patches using an encoder-decoder structure, enabling efficient and powerful pretraining for vision tasks. However, the MAE pretraining strategy is not effective for Mamba, which hinders our ability to pretrain the hybrid Mamba-Transformer backbones. Although the hybrid architecture using MAP still struggles to surpass the Transformer using MAE under the same settings, it provides a balance between computational cost and performance, as well as the potential to introduce longer contexts while maintaining scalability. This property is particularly important in the video domain and large language models, and we leave it for future work.


\noindent\textbf{Self-Supervised Visual Representation Learning.}
Self-Supervised Visual Representation Learning is a machine learning approach that enables the extraction of meaningful visual features from large amounts of unlabeled data. This methodology relies on pretext tasks, which serve as a means to learn representations without the need for explicit labels. GPT-style AR\citep{han2021pre} models predict the next part of an image or sequence given the previous parts, encouraging the model to understand the spatial or temporal dependencies within the data. MAE\citep{he2022masked} methods mask out random patches of an input image and train the model to reconstruct these masked regions. This technique encourages the model to learn contextual information and global representations. Contrastive Learning(CL)\citep{he2020momentum} techniques involve contrasting positive and negative samples to learn discriminative features. It typically involves creating pairs of positive and negative examples and training the model to distinguish between them. However, we found that existing pretraining strategies fail to fully unlock the potential of the hybrid framework, which motivated us to explore a new pretraining paradigm for hybrid Mamba-Transformer backbones. In this paper, we propose MAP to pretrain hybrid architectures. On the pure Mamba architecture, our method outperforms AR, due to MAP's ability to effectively model local features and the associations between local regions. On the pure Transformer architecture, although MAP does not surpass the performance of MAE, it still achieves comparable results. This is because, while local MAE reduces the receptive field, the autoregressive modeling between local regions enhances the ability to capture local relationships, ensuring that MAP maintains strong performance on Transformers.

\section{Polit Study}
Before exploring the strategies for pretraining hybrid frameworks, we first analyze how to properly pretrain Transformers and Mamba respectively. Analyzing the pretraining strategies for them helps us gain a deeper understanding of the key aspects of training hybrid frameworks. 
In this Section, we first conduct experiments to investigate the differences in pretraining strategies for ViT and Vim, which are representative of visual Transformers and visual Mamba.
The success of the MAE strategy on the ViT architecture is well acknowledged, while the Vim pretraining strategy remains in its early stages. We are interested in determining whether the MAE strategy is equally applicable to Vim or if the AR strategy is more suitable. To explore this, we conduct experiments on the classification task using the ImageNet-1K dataset. The results are shown in Table~\ref{tab:pilot1}.

\begin{table}[h]
    \centering
    \begin{tabular}{c|cccc}
    \hline
        Method & ViT & ViT+MAE & ViT+AR &ViT+CL\\
        \hline
        Accuracy & 82.3 & \textbf{83.6\small(+1.4)} & 82.5\small(+0.2) & 82.5(+0.2)\\
        \hline
        \hline
        Method & Vim & Vim+MAE & Vim+AR &Vim+CL \\
        \hline
        Accuracy & 81.2 & 81.4\small(+0.2) & \textbf{82.6\small(+1.4)} & 81.1\small(-0.1)\\
    \hline
    \end{tabular}
    \vspace{-3mm}
    \caption{Pilot Study. We use ViT-B and Vim-B as the default configurations. The AR strategy processes the image tokens in a row-first order, while the MAE operates according to the default settings. For contrastive learning, we only used crop and scale data augmentation and used the MoCov2 for pretraining. All experiments are conducted at a resolution of 224x224. The number of mask tokens for AR is set to 40 tokens (20\%). }
    \label{tab:pilot1}
    \vspace{-3mm}
\end{table}

Experiments show that MAE is more suitable for Transformer pretraining, while AR is better suited for Mamba pretraining.
We observe that the MAE strategy significantly enhances the performance of ViT. However, for Vim, the MAE strategy does not yield the expected improvements, while the AR strategy substantially boosts its performance. This indicates that for the ViT, applying the MAE strategy is essential to establish bidirectional associations between tokens, thereby improving performance. In contrast, for Vim, it is more important to model the continuity between preceding and succeeding tokens. Therefore, for the hybrid framework, we intuitively need a strategy that can leverage the advantages of MAE in bidirectional modeling while retaining the strengths of AR in context modeling. Since MAE's pretraining on Transformers is already well-established, benefiting from its asymmetric network design and high masking ratio, the key to AR's success on Mamba remains unclear. Therefore, in the following sections, we analyze the critical factors behind AR's success on Mamba, which will help us design strategies for the hybrid framework later.

We conducted an in-depth analysis and discovered that consistent autoregression pretraining with scanning order and proper masking ratio is the key to pretraining Mamba. This is crucial for diagnosing how to pretrain Mamba and for designing pretraining strategies for the hybrid backbone.

\begin{figure}[h]
    \centering
    \includegraphics[width=0.6\linewidth]{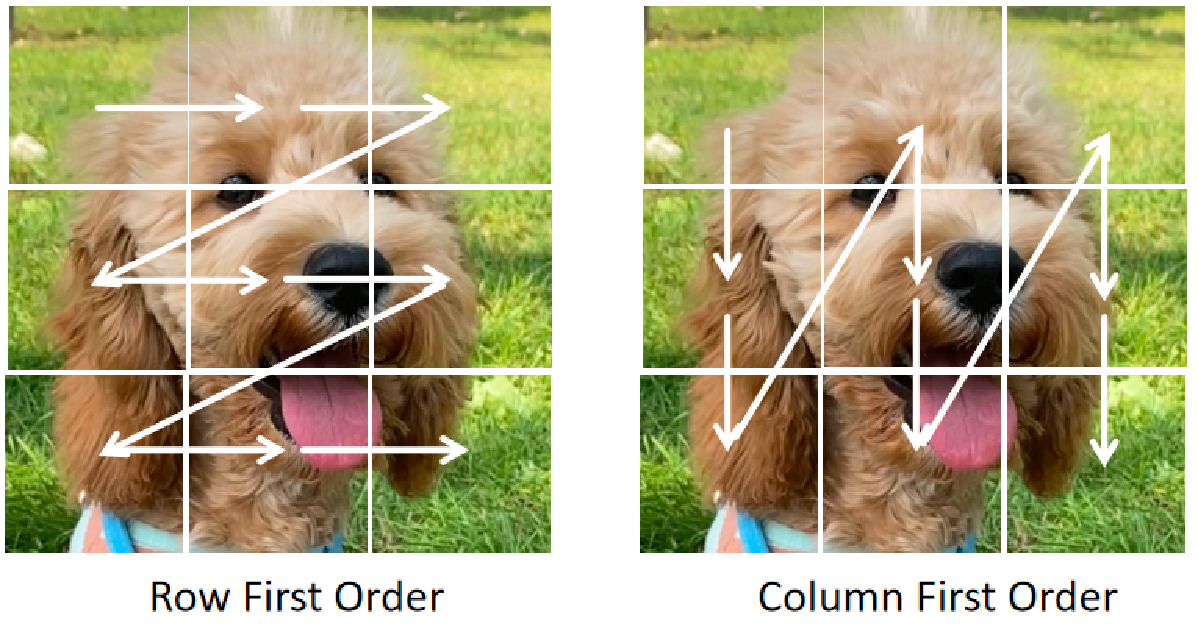}
    \vspace{-3mm}
    \caption{Different orders for AR pretraining and Mamba scanning. The row-first and column-first orders allow the network to perceive local information in different ways and sequences.}
    \label{fig:order}
    \vspace{-3mm}
\end{figure}

\noindent\textbf{Relationship between AR and Scanning Order.}
Different prediction orders can significantly impact how the model captures image features and the effectiveness of sequence generation. The analysis of the role of prediction order will help optimize AR pretraining for Vim, exploring how the model can better capture the continuity and relationships of image information under different contextual conditions. We conduct ablation studies on Vim by allowing it to perform both row-first and column-first scanning. We then pretrain it with row-first and column-first AR orders, respectively, to compare their performance. Figure~\ref{fig:order} shows different orders for AR pretraining and Mamba scanning.

\begin{table}[h]
    \centering
    \begin{tabular}{c|ccc}
    \hline
        Method & Vim\small{(R)} & Vim\small{(R)} + AR\small{(C)} & Vim\small{(R)} + AR\small(R)  \\
        \hline
        Accuracy & 79.7 & 79.9\small(+0.2) & \textbf{82.6\small(+2.9)} \\
        \hline
        \hline
        Method & Vim\small{(C)} & Vim\small{(C)} + AR\small{(C)} & Vim\small{(C)} + AR\small{(R)} \\
        \hline
        Accuracy & 79.5 & \textbf{82.5\small(+3.0)} & 79.9\small(+0.4)   \\
    \hline
    \end{tabular}
    \vspace{-3mm}
    \caption{The impact of AR order on downstream tasks. Vim\small(R) refers to Vim with row-first scanning. Vim\small(C) refers to Vim with column-first scanning. AR\small{(R)} refers to row-first autoregressive pretraining. AR\small{(C)} refers to column-first autoregressive pretraining. The results indicate that the best performance is achieved when the AR pretraining design aligns with Mamba's scanning order.}
    \label{tab:pilot2}
    \vspace{-3mm}
\end{table}

The results are shown in Table \ref{tab:pilot2}. We observe that employing a pretraining strategy consistent with the scanning order significantly enhances Vim's performance. This suggests that when designing pretraining strategies, they should be aligned with the downstream scanning order. This conclusion is also implicitly confirmed in ARM\cite{ren2024autoregressive}, as it uses cluster-based AR to pretrain cross-scanned Mamba, which is essentially a combination of row-first and column-first approaches. Unlike ARM, we experimentally verified this and provided comprehensive experimental results.

\noindent\textbf{Masking Ratio of Autoregression Pretraining.} 
Masking a single token follows the traditional AR paradigm, while masking \( n \) tokens transforms the task into an inpainting problem, as the input and output sequence lengths remain equal.  In this context, varying the auto-regressive masking ratios effectively adjusts the inpainting ratio, influencing the model's predictions beyond just the sequence length.


\begin{table}[h]
    \centering
    \begin{tabular}{c|ccc}
    \hline
         Masked tokens & 1 (0.5\%) & 20 (10\%) & 40 (20\%)  \\
        \hline
        Accuracy & 81.7 & 82.0 & \textbf{82.6} \\
        \hline
        \hline
         Masked tokens & 60 (30\%) & 100 (50\%) & 140 (70\%)\\
        \hline
        Accuracy &  82.5 & 82.2 & 81.9   \\
    \hline
    \end{tabular}
    \vspace{-3mm}
    \caption{The impact of Masking Ratio on AR pretraining. We masked 1 token (0.5\%), 20 tokens (10\%), 40 tokens (20\%), 60 tokens (30\%), 100 tokens (50\%), and 140 tokens (70\%), respectively, while also recording the results of fine-tuning on downstream tasks. The experiment shows that an appropriate masking ratio is important for autoregressive pretraining. }
    \label{tab:pilot3}
    \vspace{-5mm}
\end{table}


The results are shown in Table~\ref{tab:pilot3}. In auto-regressive pretraining, as the Masking Ratio increases, the performance of the Mamba improves. This is because a higher Masking Ratio encourages the model to learn more complex and rich feature representations, thereby enhancing its generative ability and adaptability.
However, an excessively high Masking Ratio may lead to instability during the training process. In such cases, the model may struggle to make accurate predictions due to a lack of sufficient contextual information, negatively impacting its pretraining effectiveness. Therefore, when designing auto-regressive pretraining tasks, finding an appropriate masking ratio is crucial to strike a balance between performance improvement and training stability. 

Now we have the following three conclusions to serve as a reference for designing the pretraining of hybrid backbone:
\begin{itemize}
    \item MAE is more suitable for Transformers, while AR is better suited for Mamba.
    \item For MAE pretraining of Transformers, an asymmetric structure, and an appropriate masking ratio are important.
    \item For AR pretraining of Mamba, an appropriate AR order and masking ratio are important.
\end{itemize}

Based on the above conclusions, we propose Masked Autoregressive Pretraining (MAP) to pretrain hybrid frameworks. MAP leverages AR to model the relationships between local regions and local MAE to model the internal features of local regions. It also employs a masking strategy that is crucial for both MAE and AR to enhance the network's representation capabilities. This strategy combines the strengths of MAE in local feature modeling, which is suitable for pretraining Transformers, and the strengths of AR in context modeling, which is suitable for pretraining Mamba. Effectively inheriting the advantages of both strategies, MAP improves the pretraining performance of hybrid architectures within a unified framework. In the following section, we will provide a detailed introduction to the MAP pretraining strategy.

\section{Method}
\begin{figure}[h]
    \centering
    \includegraphics[width=0.7\linewidth]{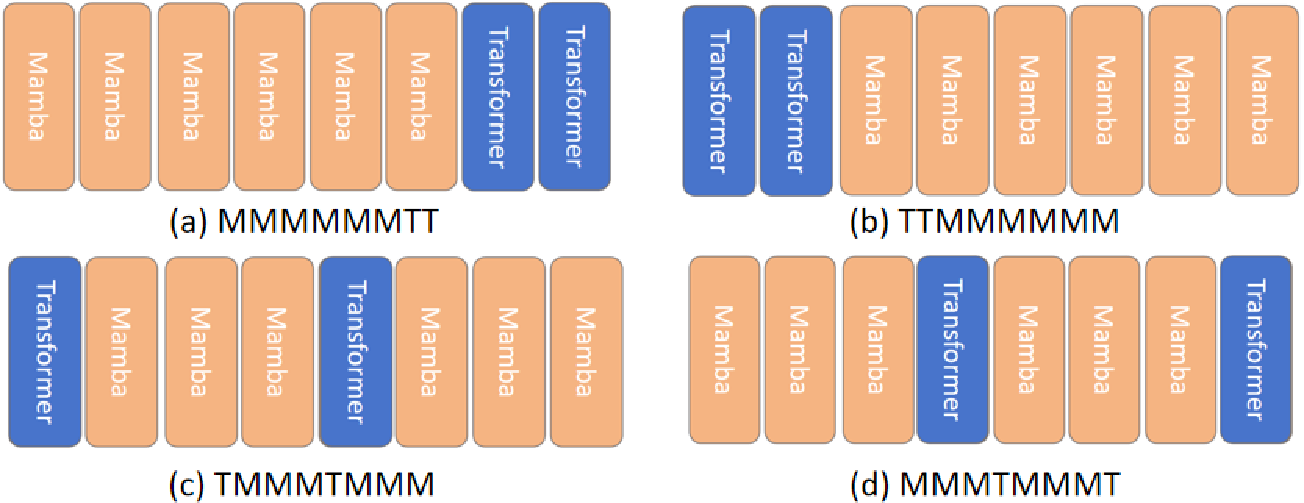}
    \vspace{-3mm}
    \caption{Different Hybrid Model Design. (d) achieves the best results and is set as default and refer to it as \textbf{HybridNet}.}
        \vspace{-5mm}
    \label{fig:hybrid}
\end{figure}

\textbf{Preliminary.}The focus of this paper is on studying how to pretrain hybrid Mamba-Transformer frameworks, rather than on designing the hybrid frameworks themselves. Therefore, we first determine a hybrid network as the default of our study. We tried a series of hybrid Mamba-Transformer vision backbones and compared their performance when trained from scratch. The results indicate that the hybrid approach using MMMTMMMT performs the best. When comparing Mamba-R* with MMMMMMTT, we found that adding a Transformer after Mamba enhances its long-context modeling capabilities, leading to improved performance. However, when comparing MMMMMMTT with TTMMMMMM, we observed that simply appending Transformers after Mamba does not fully leverage the architecture's potential. This suggests that incorporating Transformers at the beginning is crucial for extracting sufficient local features. We believe that the MMMTMMMT approach effectively balances local feature extraction and contextual modeling enhancement, making it our default configuration. We adopt the network shown in Figure~\ref{fig:hybrid}(d) as our default hybrid network and refer to it as \textbf{HybridNet}. It is worth noting that our method can also be applied to other Mamba-Transformer hybrid frameworks, but the design of the hybrid frameworks themselves is not within the scope of this paper. 

\begin{table}[h]
    \centering
    \begin{tabular}{c|ccc}
    \hline
         Design & \small{DeiT*} & \small{Mamba-R*} & \scriptsize{MMMMMMTT}  \\
        \hline
        Accuracy & 82.80 & 82.70 & 82.88 \\
        \hline
        \hline
          Design & \scriptsize{TTMMMMMM} & \scriptsize{TMMMTMMM} & \scriptsize{MMMTMMMT}\\
        \hline
        Accuracy &  82.93& 83.01 & \textbf{83.12}  \\
    \hline
    \end{tabular}
    \vspace{-3mm}
    \caption{Hybrid Design of Mamba-Transformer backbone. All experiments are trained from scratch. Mamba-R* means 24 Mamba-R\citep{wang2024mamba} Mamba layers plus 8 additional Mamba layers. DeiT* means 24 DeiT\citep{deit} Transformer layers plus 8 additional Transformer layers. MMMMMMTT represents 24 Mamba layers followed by 8 Transformer layers. TTMMMMMM represents 8 Transformer layers followed by 24 Mamba layers. TMMMTMMM represents a unit consisting of 1 Transformer layer and 3 Mamba layers, repeated 8 times. MMMTMMMT represents a unit of 3 Mamba layers followed by 1 Transformer layer, repeated 8 times.}
    \label{tab:hybrid}
    \vspace{-3mm}
\end{table}

\noindent\textbf{Overview.}
Our proposed MAP is a pre-training strategy for the hybrid Mamba-Transformer framework that can significantly improve the performance of the hybrid architecture compared to traditional MAE and AR pre-training.  Our MAP pre-training is to perform random masking on a given image and then region-wise autoregressively reconstruct its original image. Here, the tokens of each row of the image are predicted at the same time, and the tokens between rows are predicted by autoregression. 

Notably, we choose each row as a sub-region here because, as mentioned in the previous chapter, the AR order of Mamba should match the scanning order. The default scanning order for most Mamba implementations is row-first. To ensure a fairer comparison with existing methods, we treat each row as a prediction unit. After changing the scanning order of Mamba, more sophisticated clustering strategies are theoretically expected to yield better results.

As shown in Figure~\ref{fig:teaser}(c), given an image, HybridNet first maps the randomly masked image into the feature space and then uses a Transformer Decoder to decode the original image row-wise.
Consider an image \( \mathbf{I} \) which is divided into rows:  
\begin{equation}
       \mathbf{I} = \{\mathbf{r}_1, \mathbf{r}_2, \ldots, \mathbf{r}_M\}  
\end{equation}

   Each row \(\mathbf{r}_i\) consists of tokens:  
\begin{equation}
   \mathbf{r}_i = \{\mathbf{x}_{i1}, \mathbf{x}_{i2}, \ldots, \mathbf{x}_{iN}\}  
\end{equation}
  
Select a subset of tokens to mask within each row. Let \(\mathbf{M}_i \subset \{1, 2, \ldots, N\}\) denote the indices of the masked tokens in row \(\mathbf{r}_i\).  
  
For a given row \(\mathbf{r}_i\), predict all masked tokens simultaneously:  
\begin{equation}
   p(\mathbf{x}_{ij} \mid \mathbf{x}_{i, j \notin \mathbf{M}_i}, \mathbf{r}_{<i})  
\end{equation}
   where \(\mathbf{r}_{<i}\) refers to all rows preceding row \(i\).  
  
The prediction of tokens in row \(i\) depends on all previous rows and the visible tokens within the row. This can be expressed as:  
\begin{equation}
   p(\mathbf{r}_i \mid \mathbf{r}_{<i}) = \prod_{j=1}^{N} p(\mathbf{x}_{ij} \mid \mathbf{x}_{i, j \notin \mathbf{M}_i}, \mathbf{r}_{<i})  
\end{equation}
  
The overall loss function is the sum of the negative log-likelihoods of the predicted tokens:  
\begin{equation} 
   \mathcal{L} = -\sum_{i=1}^{M} \sum_{j \in \mathbf{M}_i} \log p(\mathbf{x}_{ij} \mid \mathbf{x}_{i, j \notin \mathbf{M}_i}, \mathbf{r}_{<i})  
\end{equation}

Our proposed MAP represents a general paradigm applicable to data across various domains, with 2D image data as an example. Our method can be easily extended to large language models and the fields of image video and point cloud video. Our method optimizes the synergy between Mamba and Transformer within a unified framework, allowing both models to fully leverage their strengths. In the Mamba-Transformer hybrid architecture, this approach effectively enhances the cooperation between the models, resulting in significant performance improvements. 

For HybridNet, the masking mechanism of MAP can significantly enhance the local modeling capabilities of both the Mamba and Transformer layers. The region-wise autoregressive strategy of MAP can significantly improve Mamba's ability to model context. Therefore, overall, MAP can bring about substantial performance improvements for HybridNet. For pure Mamba architecture, our method outperforms AR, due to MAP's ability to effectively model local features and the associations between local regions. On the pure Transformer architecture, although MAP does not surpass the performance of MAE, it still achieves comparable results. This is because, while local MAE reduces the receptive field, the autoregressive modeling between local regions enhances the ability to capture local relationships, ensuring that MAP maintains strong performance on Transformers.

We will introduce the specific design components of the framework. The subsequent experiments in this section are conducted using the base-sized model on the ImageNet-1K dataset.

\noindent\textbf{Masking.}
Since masking is also crucial for MAE pretraining of Transformers and AR pretraining of Mamba, masking is an important design aspect of MAP.
We experimented with different masking strategies, including random, sequential, and diagonal masking. Our experiments show that random masking delivers the best results. We attribute this to the fact that sequential and diagonal masking can hinder the Transformer's ability to establish contextual relationships. Random masking not only promotes bidirectional modeling for Transformers but also enhances Mamba's generalization and representation capabilities in sequence modeling. The random masking strategy is actually a balance, as it retains the bidirectional modeling capability of Transformers while leveraging row-wise autoregressive pretraining to train Mamba. The raster and diagonal strategies fail to fully utilize the bidirectional modeling capability of Transformers and do not enhance Mamba's ability to model local features.

Additionally, we explored the effects of different masking ratios and found that a 50\% masking ratio yielded the best results. This conclusion aligns with intuition: while MAE performs optimally on Transformers with a 75\% masking ratio, previous experiments showed that AR achieves the best results on Mamba with a 20\% ratio. Therefore, a 50\% ratio serves as a balanced number, leveraging the strengths of both paradigms. This ratio is significantly different from the conclusions drawn in MAE.

\begin{figure}[h]
    \centering
    \includegraphics[width=0.9\linewidth]{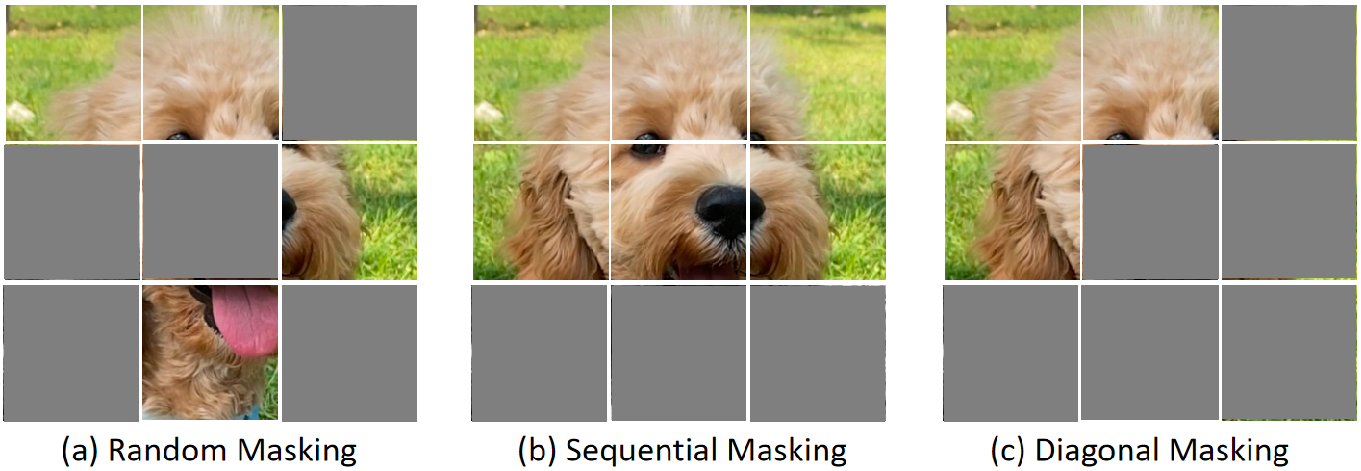}
    \vspace{-3mm}
    \caption{Different Masking Strategies. The random masking strategy produces the best results.}
    \label{fig:masking}
    \vspace{-5mm}
\end{figure}

\begin{table}[h]
    \centering
    \begin{tabular}{c|cccc}
    \hline
         Strategy & {FS} &  {Random} &   {Sequential} &  {Diagonal } \\
        \hline
        Accuracy & 83.1 & \textbf{84.9} & 84.0 & 83.8  \\
        \hline
        \hline
        Ratio &  \textbf{0\%} & \textbf{25\%} &  \textbf{50\%} &  \textbf{75\%}  \\
        \hline
        Accuracy & 83.3 & 84.5 & \textbf{84.9} & 84.2 \\
        \hline
    \end{tabular}
    \vspace{-3mm}
    \caption{Random masking with a 50\% ratio performs the best.}
    \label{tab:hybrid}
    \vspace{-5mm}
\end{table}

\noindent\textbf{MAP Transformer Decoder.} To reconstruct the original image, we utilize a masked Transformer for signal recovery. The reason for using a Transformer decoder rather than a Mamba decoder is that the Transformer decoder can reconstruct region-wise based on the encoder's features by applying a decoder mask. In contrast, the Mamba decoder, due to its unidirectional scanning nature, struggles to simultaneously reconstruct an entire local region.
Our decoder employs a distinct row-wise decoding strategy that allows autoregressive decoding of one row of tokens at a time, enhancing the network's ability to capture local features and contextual relationships among regions. Experiments show that this method significantly outperforms the original AR, MAE, and local MAE decoding strategies. Local MAR refers to limiting the receptive field of MAE within each row of the image. Notably, in the hybrid framework, local MAE performs comparably to standard MAE, emphasizing the significance of local feature learning. Our MAP method improves local feature modeling while leveraging autoregressive techniques to capture contextual relationships across regions, resulting in superior performance.

\begin{figure}[h]
    \centering
    \includegraphics[width=1\linewidth]{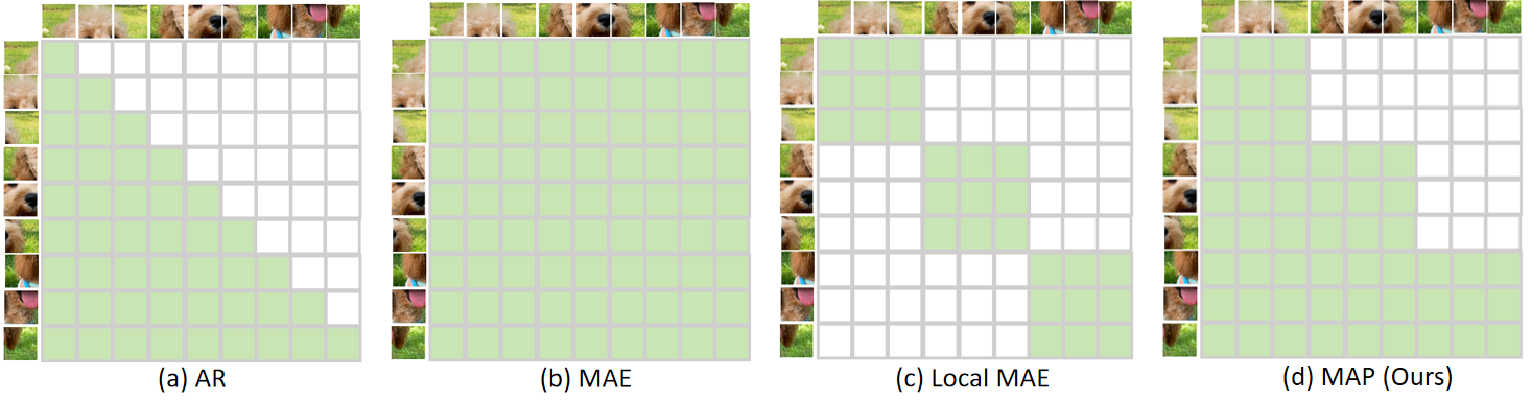}
    \vspace{-3mm}
    \caption{Different Decoder Mask. Green represents activation. White represents non-activation.}
    \vspace{-5mm}
    \label{fig:mask}
\end{figure}

\begin{table}[h]
    \centering
    \begin{tabular}{c|cccc}
    \hline
         Decoder Mask & {AR} &  {MAE} &   {local MAE} &  ours \\
        \hline
        Accuracy & 83.7 & 84.1 & 84.2 & \textbf{84.9}  \\
        \hline
    \end{tabular}
    \vspace{-3mm}
    \caption{Decoder Mask Design. Our MAP decoder strategy achieves the best results.}
    \label{tab:Decoder}
    \vspace{-4mm}
\end{table}

\textbf{Reconstruction Target.} Consistent with MAE, we reconstructed normalized original pixels as the target and employed MSE loss. Inspired by MAR\citep{li2024autoregressive} to use reconstruction output as a conditional signal for diffusion models to improve generation quality, we explored whether pretraining with diffusion loss could enhance performance. However, this approach did not yield significant improvements, suggesting that the quality of reconstructed images is not directly linked to encoder pretraining success.

\begin{table}[H]
    \centering
    \begin{tabular}{c|ccc}
    \hline
         Loss & FS &  {Diffusion Loss} &   MSE Loss (ours) \\
        \hline
        Accuracy & 83.1 & 83.3  & \textbf{84.9}  \\
        \hline
    \end{tabular}
    \vspace{-3mm}
    \caption{Reconstruction Target.}
    \label{tab:Reconstruction}
\end{table}


\section{Experiments}
\label{sec:experiments}
\subsection{2D Classification on ImageNet-1k.}
\textbf{Settings.} We pretrained on the training set of the ImageNet-1K\citep{imagenet} dataset and then fine-tuned on its classification task. We report the top-1 validation accuracy of a single 224x224 crop, and in some settings, we also report the results for a 384x384 crop. During the pretraining phase, we applied a random masking strategy with a 50\% masking ratio, using only random cropping as the data augmentation strategy. We utilized AdamW as the optimizer and trained for 1600 epochs across all settings. 
In the fine-tuning phase, we directly fine-tune for 400 epochs and report the results.

\begin{table}[h]
\centering
\resizebox{\linewidth}{!}{
\begin{tabular}{lccccc}
\toprule
Model      & Img. size & \#Params & Throughput & Mem & Acc. (\%) \\
\midrule
\multicolumn{5}{l}{\bf \textit{Pure Convolutional networks:}} \\
ResNet-50~\citep{resnet} & 224$^2$ & 25M & 2388 &6.6G & 76.2 \\
ResNet-152~\citep{resnet} & 224$^2$ & 60M & 1169 &12.5G & 78.3 \\
EfficientNet-B3~\citep{efficientnet} & 300$^2$ & 12M & 496 &19.7G & 81.6 \\
ConvNeXt-T~\citep{convnext} & 224$^2$ & 29M & 701 & 8.3G & 82.1 \\
ConvNeXt-S~\citep{convnext} & 224$^2$ & 50M & 444 &13.1G & 83.1 \\
ConvNeXt-B~\citep{convnext} & 224$^2$ & 89M & 334 &17.9G & 83.8 \\
\midrule

\multicolumn{5}{l}{\bf \textit{Pure Vision Transformers:}} \\
ViT-B/16~\citep{vit} & 224$^2$ & 86M & 284 &63.8G & 77.9 \\
ViT-L/16~\citep{vit} & 224$^2$ & 307M & 149  & - & 76.5 \\
\multicolumn{5}{l}{\bf \textit{Pretrained Vision Transformers:}} \\
ViT-B/16 + AR~\citep{vit} & 224$^2$ & 86M &  284 &63.8G & 82.5\\
ViT-B/16 + MAE~\citep{vit} & 224$^2$ & 86M &  284 &63.8G & 83.6\\
ViT-L/16 + MAE~\citep{vit} & 224$^2$ & 307M & 149 &- & 85.9 \\\rowcolor{cyan!10}
ViT-B/16 + MAP & 224$^2$ & 86M & 284 &63.8G & {83.6} \\\rowcolor{cyan!10}
ViT-L/16 + MAP & 224$^2$ & 307M & 149  &-&{86.1} \\
\midrule

\multicolumn{5}{l}{\bf \textit{Pure Mamba architecture:}} \\
Vim-T~\citep{vim} & 224$^2$ & 7M & 1165 &4.8G  & 76.1 \\
Vim-S~\citep{vim} & 224$^2$ & 26M & 612 &9.4G & 80.5 \\
MambaR-T~\citep{wang2024mamba} & 224$^2$ & 9M & 1160  &5.1G & 77.4 \\
MambaR-S~\citep{wang2024mamba} & 224$^2$ & 28M & 608 &9.9G & 81.1     \\
MambaR-B~\citep{wang2024mamba} & 224$^2$ & 99M & 315 &20.3G & 82.9     \\
MambaR-L~\citep{wang2024mamba} & 224$^2$ & 341M & 92 &55.5G & 83.2 \\
\multicolumn{5}{l}{\bf \textit{Pretrained Mamba architecture:}} \\
ARM-B (Mamba+AR)~\citep{ren2024autoregressive} &  224$^2$& 85M&325& 19.7G& 83.2  \\
ARM-L (Mamba+AR)~\citep{ren2024autoregressive}  & 224$^2$ &297M&111  & 53.1G &84.5     \\

MambaR-B+MAE &  224$^2$& 99M&315 & 20.3G& {83.1} \\

MambaR-B+AR &  224$^2$& 99M&315 & 20.3G& {83.7} \\\rowcolor{cyan!10}

MambaR-B+MAP &  224$^2$& 99M&315 & 20.3G& {84.0} \\\rowcolor{cyan!10}
   
MambaR-L+MAP  & 224$^2$ &341M&92   &55.5G&  {84.8}  \\

\midrule

\multicolumn{5}{l}{\bf \textit{Hybrid 2D convolution + Mamba:}} \\
VMamba-T~\citep{vmamba} & 224$^2$ & 31M & 464 &7.6G & 82.5 \\
VMamba-S~\citep{vmamba} & 224$^2$ & 50M & 313 &27.6G & 83.6 \\
VMamba-B~\citep{vmamba} & 224$^2$ & 89M & 246 &37.1G & 83.9 \\
\midrule

\multicolumn{5}{l}{\bf \textit{Hybrid 2Dconvolution + Mamba + Transformer architecture: (with down-sampling)}} \\
MambaVision-T~\citep{hatamizadeh2024mambavision} & 224$^2$ & 35M & 1349  &10.7G& 82.7 \\
MambaVision-S~\citep{hatamizadeh2024mambavision}  & 224$^2$ & 51M & 1058  & 36.6G& 83.3    \\
MambaVision-B~\citep{hatamizadeh2024mambavision}  & 224$^2$ & 97M & 826 &50.8G & 84.2    \\
MambaVision-L ~\citep{hatamizadeh2024mambavision} & 224$^2$ & 241M & 229  &78.6G & 85.3 \\\rowcolor{cyan!10}
MambaVision-B+MAP~\citep{hatamizadeh2024mambavision}  & 224$^2$ & 97M & 826 &50.8G & {84.9}    \\\rowcolor{cyan!10}
MambaVision-L+MAP ~\citep{hatamizadeh2024mambavision} & 224$^2$ & 241M & 229  &78.6G & {86.4} \\
\midrule

\multicolumn{5}{l}{\bf \textit{Hybrid Mamba + Transformer architecture: (without down-sampling)}} \\
HybridNet-T & 224$^2$ & 12M &  910 &7.6G& 77.7 \\
HybridNet-S & 224$^2$ & 37M & 512 &14.6G & 81.3     \\
HybridNet-B & 224$^2$ & 128M &  244 &30.0G & 83.1    \\
          & 384$^2$ & 128M&  244& 76.1G &  84.5   \\
HybridNet-L & 224$^2$ & 443M & 63  &78.3G & 83.2  \\
          & 384$^2$ & 443M &  63 &- & 84.6  \\
\multicolumn{5}{l}{\bf \textit{Pretrained Hybrid architecture:}} \\\rowcolor{cyan!10}
HybridNet-T + MAP & 224$^2$ & 12M &   910  &7.6G&  {78.6}\\\rowcolor{cyan!10}
HybridNet-S + MAP & 224$^2$ & 37M & 512 &14.6G &  {82.5}  \\

HybridNet-B + MAE & 224$^2$ & 128M   & 244&30.0G & {83.9}  \\
HybridNet-B + AR & 224$^2$ & 128M   & 244&30.0G & {83.8}  \\
HybridNet-B + CL & 224$^2$ & 128M   & 244&30.0G & {83.1}  \\\rowcolor{cyan!10}

HybridNet-B + MAP & 224$^2$ & 128M   & 244&30.0G & {84.9}  \\\rowcolor{cyan!10}
 & 384$^2$ &128M &  244& 76.1G&  {85.5}  \\\rowcolor{cyan!10}

HybridNet-L + MAP & 224$^2$ & 443M & 63  &78.3G & {85.0}\\\rowcolor{cyan!10}
 & 384$^2$ & 443M & 63 &-  & {86.2}\\
\bottomrule
\end{tabular}

}

\vspace{-3mm}
\caption{ImageNet-1k classification results. The throughput is computed on an A100 GPU. The memory overhead is measured with a batch size of 128. Our results are highlighted in \colorbox{cyan!10}{blue}. }
\vspace{-3mm}
\label{tab:in1k}
\end{table}

\textbf{Results.} Results are shown in Table~\ref{tab:in1k}. The results indicate that the hybrid framework achieves a balance between performance and computational overhead. 
The focus of this paper is not on designing the hybrid network but on investigating its pretraining methods. Our proposed pretraining method MAP significantly enhances the performance of the hybrid Mamba-Transformer framework. Comparing the performance of MAP, MAE, AR, and CL on HybridNet-B, it is evident that MAP significantly improves performance. This indicates that compared to MAE, the local MAE in MAP, which enhances local modeling and adopts an autoregressive approach to strengthen the associations between local regions, is crucial for hybrid networks. Compared to AR, the masking mechanism and the local autoregressive strategy in MAP contribute to performance gains.
Additionally, we verified that our MAP method can improve the performance of the pure Mamba framework and the pure Transformer backbone. By comparing AR, MAE, and MAP pretraining results on MambaR-B, we can see that MAE offers limited improvement for pure Mamba. AR, benefiting from its autoregressive modeling of context, shows improved performance. Compared to AR, the masking mechanism and local autoregressive decoding strategy in MAP enhance Mamba's ability to model local features, further improving performance. By comparing the performance of AR, MAE, and MAP on ViT, we find that on ViT-B, MAP can achieve the same results as MAE. This is because, although local MAE reduces the receptive field compared to the original MAE, region-wise autoregressive modeling not only enhances local feature modeling but also strengthens the modeling of correlations between local regions, preventing a decline in performance. When comparing MAE and MAP on ViT-L, the advantage of region-wise autoregressive modeling in MAP becomes apparent in larger-scale models, where it outperforms MAE. This also suggests that autoregressive modeling has advantages in larger models, which is consistent with the scaling law phenomenon observed in LLMs. In addition, by comparing the results of HybridNet-B/L at 224 and 384 resolutions, we can see that at the longer sequence length of 384 resolution, our method further achieved improvements of 0.6 and 0.8, respectively. This indicates that the longer context modeling allowed by the Mamba layer can indeed lead to performance gains. Comparing the results of MambaVision-B/L before and after MAP pretraining shows that our MAP method not only benefits the HybridNet used as an example but also brings significant improvements to other existing hybrid Mamba-Transformer network frameworks.

\subsection{Transfer Learning Experiments}
\textbf{Semantic Segmentation}
We conduct experiments for semantic segmentation on the ADE20K~\cite{zhou2017scene} and use UperNet~\cite{xiao2018unified} as the segmentation framework followed by Vim. All settings are consistent with Vim. When comparing HybridNet-S with Vim-S, we can see that introducing a Transformer significantly enhances the performance of dense prediction tasks. This highlights the value of hybrid networks, which maintain satisfactory performance when introducing longer contexts at 512x512 resolution. Comparing the results of HybridNet-S before and after MAP pretraining, we can see that our pretraining method also brings significant performance improvements in segmentation tasks.

\noindent\textbf{Object Detection and Instance Segmentation}
We conduct experiments for object detection and instance segmentation on the COCO 2017 dataset~\cite{lin2014microsoft} and use ViTDet~\cite{xiao2018unified} as the basic framework followed by Vim. All settings are consistent with Vim. The results indicate that, while simply inserting a Transformer into Vim does not bring about significant improvements, the performance of HybridNet-Ti, when fine-tuned after initialization with MAP pretraining, shows a notable improvement.

\begin{table}[t]
\centering
\scriptsize
\addtolength{\tabcolsep}{-1pt}
\begin{tabular}{l c | c  c  | c }
\toprule
Method &  Backbone & \begin{tabular}[c]{@{}c@{}}image \\ size\end{tabular} & \#param.  & \begin{tabular}[c]{@{}c@{}}$val$ \\ mIoU\end{tabular} \\
\toprule
DeepLab v3+ & ResNet-101&$512^{2}$ & 63M  & 44.1\\ %
UperNet  & ResNet-50 &$512^{2}$ & 67M  & 41.2 \\ %
UperNet  & ResNet-101 &$512^{2}$ & 86M  & 44.9 \\ %
\midrule
UperNet  & DeiT-Ti&$512^{2}$ & 11M  & 39.2  \\ %
UperNet  & DeiT-S&$512^{2}$ & 43M  & 44.0 \\ %
\midrule
\rblue
UperNet  & Vim-Ti &$512^{2}$ & 13M & 41.0 \\
\rblue
UperNet  & Vim-S &$512^{2}$ & 46M & 44.9 \\
UperNet  & HybridNet-S &$512^{2}$ & 69M & 45.6 \\\rowcolor{cyan!10}
UperNet  & HybridNet-S+MAP &$512^{2}$ & 69M & 46.9 \\
\bottomrule
\end{tabular}
\vspace{-3mm}
\caption{Results of semantic segmentation on the ADE20K $val$ set. }
\vspace{-5mm}
\label{tab:segcomp}
\end{table}

\begin{table}[t]
\centering
\scriptsize
\addtolength{\tabcolsep}{-1pt}
\begin{tabular}{m{80pt} | m{40pt} | m{40pt}}
\toprule
Backbone & AP$^{\text{box}}$ & AP$^{\text{mask}}$ \\
\midrule
DeiT-Ti & 44.4 & 38.1 \\
\rblue
Vim-Ti & 45.7 & 39.2 \\
HybridNet-Ti & 45.9 & 39.2 \\
\rowcolor{cyan!10}
HybridNet-Ti+MAP & 46.4 & 39.8 \\
\bottomrule
\end{tabular}
\vspace{-3mm}
\caption{Results of object detection and instance segmentation on the COCO $val$ set using Cascade Mask R-CNN~\cite{cai2019cascade} framework. }
\vspace{-5mm}
\label{tab:detcomp}
\end{table}

\begin{table}[t]
\renewcommand{\arraystretch}{1.00}
\begin{center}
\scriptsize
\resizebox{\columnwidth}{!}{
    \begin{tabular}{lcccc}
    \toprule
    Method & \texttt{PT} & \#P $\downarrow$ & \#F $\downarrow$ & ModelNet40 \\
    \midrule
    \multicolumn{5}{c}{\textit{Supervised Learning Only: Dedicated Architectures}}\\
    \midrule
    PointNet\tiny\citep{PointNet} & $\times$ & 3.5 & 0.5 & 89.2 \\
    PointNet++\tiny\citep{PointNet++} & $\times$ & 1.5 & 1.7 & 90.7 \\
    DGCNN\tiny\citep{DGCNN} & $\times$ & 1.8 & 2.4 & 92.9 \\
    PointCNN\tiny\citep{PointCNN} & $\times$ & 0.6 & - & 92.2 \\
    DRNet \tiny\citep{DRNet} & $\times$ & - & - & 93.1 \\
    SimpleView\tiny\citep{SimpleView} & $\times$ & - & - & 93.9 \\
    GBNet\tiny\citep{GBNet} & $\times$ & 8.8 & - & 93.8 \\
    PRA-Ne\tiny\citep{PRANet} & $\times$ & - & 2.3 & 93.7 \\
    MVTN\tiny\citep{MVTN} & $\times$ & 11.2 & 43.7 & 93.8 \\
    PointMLP\tiny\citep{PointMLP} & $\times$ & 12.6 & 31.4 & 94.5 \\
    PointNeXt\tiny\citep{PointNeXt} & $\times$ & 1.4 & 3.6 & 94.0 \\
    P2P-HorNet\tiny\citep{P2P} & $\checkmark$ & - & 34.6 & 94.0 \\
    DeLA\tiny\citep{DeLA} & $\times$ & 5.3 & 1.5 & 94.0 \\
    \midrule
    \multicolumn{5}{c}{\textit{Supervised Learning Only: Transformer or Mamba-based Models}}\\
    \midrule
    Transformer & $\times$ & 22.1 & 4.8 & 91.4 \\
    PCT\tiny\citep{PCT} & $\times$ & 2.9 & 2.3 & 93.2 \\
    PointMamba\tiny\citep{liang2024pointmamba} & $\times$ & 12.3 & 3.6 & - \\
    PCM\tiny\citep{PCM24} & $\times$ & 34.2 & 45.0 & 93.4$\pm$0.2 \\
    SPoTr\tiny\citep{SPoTr} & $\times$ & 1.7 & 10.8 & - \\
    PointConT\tiny\citep{PointConT} & $\times$ & - & - & 93.5 \\
    Mamba3d w/o vot.\tiny\citep{han2024mamba3d} & $\times$ & 16.9 & 3.9 & 93.4 \\
    Mamba3d w/ vot.\tiny\citep{han2024mamba3d} & $\times$ & 16.9 & 3.9 & 94.1 \\
    HybridNet w/o vot. & $\times$ & 19.3 & 4.4 & 93.5 \\
    HybridNet w/ vot. & $\times$ & 19.3 & 4.4 & 94.3 \\
    \midrule
    \multicolumn{5}{c}{\textit{With Self-supervised pretraining}}\\
    \midrule
    Transformer & \textit{OcCo\tiny\citep{wang2021unsupervised}} & 22.1 & 4.8 & 92.1 \\
    Point-BERT & \textit{IDPT\tiny\citep{zha2023instance}} & 22.1+1.7$^\dagger$ & 4.8 & 93.4 \\
    MaskPoint & \textit{MaskPoint\tiny\citep{liu2022masked}} & 22.1 & 4.8 & 93.8 \\
    PointMamba & \textit{Point-MAE\tiny\citep{pang2022masked}} & 12.3 & 3.6 & - \\
    Point-MAE & \textit{IDPT\tiny\citep{zha2023instance}} & 22.1+1.7$^\dagger$ & 4.8 & 94.4 \\
    Point-M2AE & \textit{Point-M2AE\tiny\citep{zhang2022point}} & 15.3 & 3.6 & 94.0 \\
    Mamba3d w/o vot. & \textit{Point-BERT\tiny\citep{yu2022point}} & 16.9 & 3.9 & 94.4 \\
    Point-MAE & \textit{Point-MAE\tiny\citep{pang2022masked}} & 22.1 & 4.8 & 93.8 \\
    Mamba3d w/o vot. & \textit{Point-MAE\tiny\citep{pang2022masked}} & 16.9 & 3.9 & 94.7 \\
    Mamba3d w/ vot. & \textit{Point-MAE\tiny\citep{pang2022masked}} & 16.9 & 3.9 & 95.4 \\
    \rowcolor{cyan!10}
    Mamba3d w/o vot. & \textit{MAP} & 16.9 & 3.9 & 95.1 \\
    \rowcolor{cyan!10}
    Mamba3d w/ vot. & \textit{MAP} & 16.9 & 3.9 & 95.6 \\
    \rowcolor{cyan!10}
    HybridNet w/o vot. & \textit{MAP} & 19.3 & 4.4 & 95.4 \\
    \rowcolor{cyan!10}
    HybridNet w/ vot. & \textit{MAP} & 19.3 & 4.4 & 95.9 \\
    \bottomrule
    \end{tabular}
}
\vspace{-3mm}
\caption{Results on 3D ModelNet classification tasks.}
\vspace{-8mm}
\end{center}

\end{table}

\subsection{3D Classification Experiments}
\textbf{Settings.} To verify that our method can be applied to other data formats, we transferred the MAP pretraining to 3D data for experimentation. We pretrained using the ShapeNet\citep{ShapeNet15} dataset, employing random rotation and translation scaling as data augmentation techniques. Each point cloud consists of 1024 points and is divided into 64 patches, with each patch containing 32 points. Since point clouds are unordered, the concept of rows does not apply here; instead, we randomly generate 32 patches each time and complete the reconstruction process in an autoregressive manner. Similar to Mamba3D\cite{han2024mamba3d}, we did not adopt any special sorting strategies but ensured that the order of pretraining matches that of the actual Mamba scans.
We conducted pretraining on both the hybrid framework and the original Mamba3D to validate their performance advantages. 
During pretraining and downstream fine-tuning, we employed the AdamW optimizer with a cosine decay strategy for 300 epochs. 

\noindent\textbf{Results.}The experiments demonstrate that our method also can improve the performance on 3D tasks. This suggests that our approach can be easily adapted to other domains. Comparing HybridNet and Mamba3d at the backbone network level, it is evident that the performance improvement solely from inserting a Transformer is limited. By comparing the results of HybridNet before and after MAP pretraining, we can see that our MAP pretraining brings about a significant improvement. This further validates the effectiveness of our proposed pretraining method. More experiments on 3D datasets can be found in the Supplementary Material.

\section{Conclusion}
\label{sec:conclusion}
In this paper, we propose Masked Autoregressive Pretraining, a pretraining framework suitable for hybrid Mamba-Transformer models. It integrates the advantages of MAE and AR, retaining their core designs to model local features while strengthening the modeling of contextual correlations. Extensive experiments have validated the effectiveness of our method and the rationality of its component design.


\clearpage
{\small
\bibliographystyle{ieeenat_fullname}
\bibliography{main}

\begin{thebibliography}{63}
\providecommand{\natexlab}[1]{#1}
\providecommand{\url}[1]{\texttt{#1}}
\expandafter\ifx\csname urlstyle\endcsname\relax
  \providecommand{\doi}[1]{doi: #1}\else
  \providecommand{\doi}{doi: \begingroup \urlstyle{rm}\Url}\fi

\bibitem[Afham et~al.(2022)Afham, Dissanayake, Dissanayake, Dharmasiri, Thilakarathna, and Rodrigo]{CrossPoint22}
Mohamed Afham, Isuru Dissanayake, Dinithi Dissanayake, Amaya Dharmasiri, Kanchana Thilakarathna, and Ranga Rodrigo.
\newblock Crosspoint: Self-supervised cross-modal contrastive learning for 3d point cloud understanding.
\newblock In \emph{IEEE/CVF Conf. Comput. Vis. Pattern Recog. (CVPR)}, 2022.

\bibitem[Cai and Vasconcelos(2019)]{cai2019cascade}
Zhaowei Cai and Nuno Vasconcelos.
\newblock Cascade r-cnn: High quality object detection and instance segmentation.
\newblock \emph{IEEE transactions on pattern analysis and machine intelligence}, 43\penalty0 (5):\penalty0 1483--1498, 2019.

\bibitem[Chang et~al.(2015)Chang, Funkhouser, Guibas, Hanrahan, Huang, Li, Savarese, Savva, Song, Su, Xiao, Yi, and Yu]{ShapeNet15}
Angel~X. Chang, Thomas~A. Funkhouser, Leonidas~J. Guibas, Pat Hanrahan, Qi{-}Xing Huang, Zimo Li, Silvio Savarese, Manolis Savva, Shuran Song, Hao Su, Jianxiong Xiao, Li Yi, and Fisher Yu.
\newblock Shapenet: An information-rich 3d model repository.
\newblock \emph{CoRR}, abs/1512.03012, 2015.

\bibitem[Chen et~al.(2023)Chen, Xia, Zang, Wang, and Li]{DeLA}
Binjie Chen, Yunzhou Xia, Yu Zang, Cheng Wang, and Jonathan Li.
\newblock Decoupled local aggregation for point cloud learning.
\newblock \emph{arXiv preprint arXiv:2308.16532}, 2023.

\bibitem[Chen et~al.(2024)Chen, Niu, Lu, Meng, and Zhou]{chen2024maskmamba}
Wenchao Chen, Liqiang Niu, Ziyao Lu, Fandong Meng, and Jie Zhou.
\newblock Maskmamba: A hybrid mamba-transformer model for masked image generation.
\newblock \emph{arXiv preprint arXiv:2409.19937}, 2024.

\bibitem[Cheng et~al.(2021)Cheng, Chen, He, Liu, and Bai]{PRANet}
Silin Cheng, Xiwu Chen, Xinwei He, Zhe Liu, and Xiang Bai.
\newblock Pra-net: Point relation-aware network for 3d point cloud analysis.
\newblock \emph{IEEE Trans. Image Process. (TIP)}, 30:\penalty0 4436--4448, 2021.

\bibitem[Deng et~al.(2009)Deng, Dong, Socher, Li, Li, and Fei-Fei]{imagenet}
Jia Deng, Wei Dong, Richard Socher, Li-Jia Li, Kai Li, and Li Fei-Fei.
\newblock {ImageNet}: A large-scale hierarchical image database.
\newblock In \emph{CVPR}, 2009.

\bibitem[Dosovitskiy et~al.(2021{\natexlab{a}})Dosovitskiy, Beyer, Kolesnikov, Weissenborn, Zhai, Unterthiner, Dehghani, Minderer, Heigold, Gelly, Uszkoreit, and Houlsby]{dosovitskiy2020image}
Alexey Dosovitskiy, Lucas Beyer, Alexander Kolesnikov, Dirk Weissenborn, Xiaohua Zhai, Thomas Unterthiner, Mostafa Dehghani, Matthias Minderer, Georg Heigold, Sylvain Gelly, Jakob Uszkoreit, and Neil Houlsby.
\newblock An image is worth 16x16 words: Transformers for image recognition at scale.
\newblock In \emph{ICLR}, 2021{\natexlab{a}}.

\bibitem[Dosovitskiy et~al.(2021{\natexlab{b}})Dosovitskiy, Beyer, Kolesnikov, Weissenborn, Zhai, Unterthiner, Dehghani, Minderer, Heigold, Gelly, Uszkoreit, and Houlsby]{vit}
Alexey Dosovitskiy, Lucas Beyer, Alexander Kolesnikov, Dirk Weissenborn, Xiaohua Zhai, Thomas Unterthiner, Mostafa Dehghani, Matthias Minderer, Georg Heigold, Sylvain Gelly, Jakob Uszkoreit, and Neil Houlsby.
\newblock An image is worth 16x16 words: Transformers for image recognition at scale.
\newblock In \emph{Int. Conf. Learn. Represent. (ICLR)}, 2021{\natexlab{b}}.

\bibitem[Goyal et~al.(2021)Goyal, Law, Liu, Newell, and Deng]{SimpleView}
Ankit Goyal, Hei Law, Bowei Liu, Alejandro Newell, and Jia Deng.
\newblock Revisiting point cloud shape classification with a simple and effective baseline.
\newblock In \emph{Proc. Int. Conf. Mach. Learn. (ICML)}, pages 3809--3820. {PMLR}, 2021.

\bibitem[Gu and Dao(2023)]{gu2023mamba}
Albert Gu and Tri Dao.
\newblock Mamba: Linear-time sequence modeling with selective state spaces.
\newblock \emph{arXiv preprint arXiv:2312.00752}, 2023.

\bibitem[Guo et~al.(2021)Guo, Cai, Liu, Mu, Martin, and Hu]{PCT}
Meng{-}Hao Guo, Junxiong Cai, Zheng{-}Ning Liu, Tai{-}Jiang Mu, Ralph~R. Martin, and Shi{-}Min Hu.
\newblock {PCT:} point cloud transformer.
\newblock \emph{Comput. Vis. Media}, 7\penalty0 (2):\penalty0 187--199, 2021.

\bibitem[Hamdi et~al.(2021)Hamdi, Giancola, and Ghanem]{MVTN}
Abdullah Hamdi, Silvio Giancola, and Bernard Ghanem.
\newblock {MVTN:} multi-view transformation network for 3d shape recognition.
\newblock In \emph{Int. Conf. Comput. Vis. (ICCV)}, pages 1--11. {IEEE}, 2021.

\bibitem[Han et~al.(2021)Han, Zhang, Ding, Gu, Liu, Huo, Qiu, Yao, Zhang, Zhang, et~al.]{han2021pre}
Xu Han, Zhengyan Zhang, Ning Ding, Yuxian Gu, Xiao Liu, Yuqi Huo, Jiezhong Qiu, Yuan Yao, Ao Zhang, Liang Zhang, et~al.
\newblock Pre-trained models: Past, present and future.
\newblock \emph{AI Open}, 2:\penalty0 225--250, 2021.

\bibitem[Han et~al.(2024)Han, Tang, Wang, and Li]{han2024mamba3d}
Xu Han, Yuan Tang, Zhaoxuan Wang, and Xianzhi Li.
\newblock Mamba3d: Enhancing local features for 3d point cloud analysis via state space model.
\newblock \emph{arXiv preprint arXiv:2404.14966}, 2024.

\bibitem[Hatamizadeh and Kautz(2024)]{hatamizadeh2024mambavision}
Ali Hatamizadeh and Jan Kautz.
\newblock Mambavision: A hybrid mamba-transformer vision backbone.
\newblock \emph{arXiv preprint arXiv:2407.08083}, 2024.

\bibitem[He et~al.(2016)He, Zhang, Ren, and Sun]{resnet}
Kaiming He, Xiangyu Zhang, Shaoqing Ren, and Jian Sun.
\newblock Deep residual learning for image recognition.
\newblock In \emph{CVPR}, 2016.

\bibitem[He et~al.(2020)He, Fan, Wu, Xie, and Girshick]{he2020momentum}
Kaiming He, Haoqi Fan, Yuxin Wu, Saining Xie, and Ross Girshick.
\newblock Momentum contrast for unsupervised visual representation learning.
\newblock In \emph{Proceedings of the IEEE/CVF conference on computer vision and pattern recognition}, pages 9729--9738, 2020.

\bibitem[He et~al.(2022)He, Chen, Xie, Li, Doll{\'a}r, and Girshick]{he2022masked}
Kaiming He, Xinlei Chen, Saining Xie, Yanghao Li, Piotr Doll{\'a}r, and Ross Girshick.
\newblock Masked autoencoders are scalable vision learners.
\newblock In \emph{Proceedings of the IEEE/CVF conference on computer vision and pattern recognition}, pages 16000--16009, 2022.

\bibitem[Li et~al.(2024)Li, Tian, Li, Deng, and He]{li2024autoregressive}
Tianhong Li, Yonglong Tian, He Li, Mingyang Deng, and Kaiming He.
\newblock Autoregressive image generation without vector quantization.
\newblock \emph{arXiv preprint arXiv:2406.11838}, 2024.

\bibitem[Li et~al.(2018)Li, Bu, Sun, Wu, Di, and Chen]{PointCNN}
Yangyan Li, Rui Bu, Mingchao Sun, Wei Wu, Xinhan Di, and Baoquan Chen.
\newblock Pointcnn: Convolution on x-transformed points.
\newblock In \emph{Adv. Neural Inform. Process. Syst. (NeurIPS)}, pages 828--838, 2018.

\bibitem[Liang et~al.(2024{\natexlab{a}})Liang, Zhou, Wang, Zhu, Xu, Zou, Ye, and Bai]{PointMamba}
Dingkang Liang, Xin Zhou, Xinyu Wang, Xingkui Zhu, Wei Xu, Zhikang Zou, Xiaoqing Ye, and Xiang Bai.
\newblock Pointmamba: A simple state space model for point cloud analysis.
\newblock \emph{arXiv preprint arXiv:2402.10739}, 2024{\natexlab{a}}.

\bibitem[Liang et~al.(2024{\natexlab{b}})Liang, Zhou, Xu, Zhu, Zou, Ye, Tan, and Bai]{liang2024pointmamba}
Dingkang Liang, Xin Zhou, Wei Xu, Xingkui Zhu, Zhikang Zou, Xiaoqing Ye, Xiao Tan, and Xiang Bai.
\newblock Pointmamba: A simple state space model for point cloud analysis.
\newblock \emph{arXiv preprint arXiv:2402.10739}, 2024{\natexlab{b}}.

\bibitem[Lieber et~al.(2024)Lieber, Lenz, Bata, Cohen, Osin, Dalmedigos, Safahi, Meirom, Belinkov, Shalev-Shwartz, et~al.]{lieber2024jamba}
Opher Lieber, Barak Lenz, Hofit Bata, Gal Cohen, Jhonathan Osin, Itay Dalmedigos, Erez Safahi, Shaked Meirom, Yonatan Belinkov, Shai Shalev-Shwartz, et~al.
\newblock Jamba: A hybrid transformer-mamba language model.
\newblock \emph{arXiv preprint arXiv:2403.19887}, 2024.

\bibitem[Lin et~al.(2014)Lin, Maire, Belongie, Hays, Perona, Ramanan, Doll{\'a}r, and Zitnick]{lin2014microsoft}
Tsung-Yi Lin, Michael Maire, Serge Belongie, James Hays, Pietro Perona, Deva Ramanan, Piotr Doll{\'a}r, and C~Lawrence Zitnick.
\newblock Microsoft coco: Common objects in context.
\newblock In \emph{Computer Vision--ECCV 2014: 13th European Conference, Zurich, Switzerland, September 6-12, 2014, Proceedings, Part V 13}, pages 740--755. Springer, 2014.

\bibitem[Liu et~al.(2022{\natexlab{a}})Liu, Cai, and Lee]{MaskPoint}
Haotian Liu, Mu Cai, and Yong~Jae Lee.
\newblock Masked discrimination for self-supervised learning on point clouds.
\newblock In \emph{Eur. Conf. Comput. Vis. (ECCV)}, 2022{\natexlab{a}}.

\bibitem[Liu et~al.(2022{\natexlab{b}})Liu, Cai, and Lee]{liu2022masked}
Haotian Liu, Mu Cai, and Yong~Jae Lee.
\newblock Masked discrimination for self-supervised learning on point clouds.
\newblock In \emph{Proc. of European Conference on Computer Vision}, 2022{\natexlab{b}}.

\bibitem[Liu et~al.(2023)Liu, Tian, Lv, Li, and Wang]{PointConT}
Yahui Liu, Bin Tian, Yisheng Lv, Lingxi Li, and Fei-Yue Wang.
\newblock Point cloud classification using content-based transformer via clustering in feature space.
\newblock \emph{IEEE/CAA Journal of Automatica Sinica}, 2023.

\bibitem[Liu et~al.(2024)Liu, Tian, Zhao, Yu, Xie, Wang, Ye, and Liu]{vmamba}
Yue Liu, Yunjie Tian, Yuzhong Zhao, Hongtian Yu, Lingxi Xie, Yaowei Wang, Qixiang Ye, and Yunfan Liu.
\newblock Vmamba: Visual state space model.
\newblock \emph{arXiv preprint arXiv:2401.10166}, 2024.

\bibitem[Liu et~al.(2021)Liu, Lin, Cao, Hu, Wei, Zhang, Lin, and Guo]{liu2021swin}
Ze Liu, Yutong Lin, Yue Cao, Han Hu, Yixuan Wei, Zheng Zhang, Stephen Lin, and Baining Guo.
\newblock Swin transformer: Hierarchical vision transformer using shifted windows.
\newblock In \emph{Proceedings of the IEEE/CVF international conference on computer vision}, pages 10012--10022, 2021.

\bibitem[Liu et~al.(2022{\natexlab{c}})Liu, Mao, Wu, Feichtenhofer, Darrell, and Xie]{convnext}
Zhuang Liu, Hanzi Mao, Chao-Yuan Wu, Christoph Feichtenhofer, Trevor Darrell, and Saining Xie.
\newblock A convnet for the 2020s.
\newblock In \emph{CVPR}, 2022{\natexlab{c}}.

\bibitem[Ma et~al.(2022)Ma, Qin, You, Ran, and Fu]{PointMLP}
Xu Ma, Can Qin, Haoxuan You, Haoxi Ran, and Yun Fu.
\newblock Rethinking network design and local geometry in point cloud: {A} simple residual {MLP} framework.
\newblock In \emph{Int. Conf. Learn. Represent. (ICLR)}. OpenReview.net, 2022.

\bibitem[Pang et~al.(2022{\natexlab{a}})Pang, Wang, Tay, Liu, Tian, and Yuan]{pang2022masked}
Yatian Pang, Wenxiao Wang, Francis~EH Tay, Wei Liu, Yonghong Tian, and Li Yuan.
\newblock Masked autoencoders for point cloud self-supervised learning.
\newblock In \emph{Proc. of European Conference on Computer Vision}, 2022{\natexlab{a}}.

\bibitem[Pang et~al.(2022{\natexlab{b}})Pang, Wang, Tay, Liu, Tian, and Yuan]{PointMAE}
Yatian Pang, Wenxiao Wang, Francis E.~H. Tay, Wei Liu, Yonghong Tian, and Li Yuan.
\newblock Masked autoencoders for point cloud self-supervised learning.
\newblock In \emph{Eur. Conf. Comput. Vis. (ECCV)}, 2022{\natexlab{b}}.

\bibitem[Park et~al.(2023)Park, Lee, Kim, Xiong, and Kim]{SPoTr}
Jinyoung Park, Sanghyeok Lee, Sihyeon Kim, Yunyang Xiong, and Hyunwoo~J Kim.
\newblock Self-positioning point-based transformer for point cloud understanding.
\newblock In \emph{IEEE/CVF Conf. Comput. Vis. Pattern Recog. (CVPR)}, pages 21814--21823, 2023.

\bibitem[Qi et~al.(2017{\natexlab{a}})Qi, Su, Mo, and Guibas]{PointNet}
Charles~Ruizhongtai Qi, Hao Su, Kaichun Mo, and Leonidas~J. Guibas.
\newblock Pointnet: Deep learning on point sets for 3d classification and segmentation.
\newblock In \emph{IEEE/CVF Conf. Comput. Vis. Pattern Recog. (CVPR)}, pages 77--85, 2017{\natexlab{a}}.

\bibitem[Qi et~al.(2017{\natexlab{b}})Qi, Yi, Su, and Guibas]{PointNet++}
Charles~Ruizhongtai Qi, Li Yi, Hao Su, and Leonidas~J. Guibas.
\newblock Pointnet++: Deep hierarchical feature learning on point sets in a metric space.
\newblock In \emph{Adv. Neural Inform. Process. Syst. (NeurIPS)}, pages 5099--5108, 2017{\natexlab{b}}.

\bibitem[Qian et~al.(2022)Qian, Li, Peng, Mai, Hammoud, Elhoseiny, and Ghanem]{PointNeXt}
Guocheng Qian, Yuchen Li, Houwen Peng, Jinjie Mai, Hasan Abed Al~Kader Hammoud, Mohamed Elhoseiny, and Bernard Ghanem.
\newblock Pointnext: Revisiting pointnet++ with improved training and scaling strategies.
\newblock In \emph{Adv. Neural Inform. Process. Syst. (NeurIPS)}, 2022.

\bibitem[Qiu et~al.(2021)Qiu, Anwar, and Barnes]{DRNet}
Shi Qiu, Saeed Anwar, and Nick Barnes.
\newblock Dense-resolution network for point cloud classification and segmentation.
\newblock In \emph{IEEE Winter Conf. Appl. Comput. Vis. (WACV)}, pages 3812--3821, 2021.

\bibitem[Qiu et~al.(2022)Qiu, Anwar, and Barnes]{GBNet}
Shi Qiu, Saeed Anwar, and Nick Barnes.
\newblock Geometric back-projection network for point cloud classification.
\newblock \emph{IEEE Trans. Multimedia (TMM)}, 24:\penalty0 1943--1955, 2022.

\bibitem[Ren et~al.(2024)Ren, Li, Tu, Wang, Shu, Zhang, Mei, Yang, Wang, Wang, et~al.]{ren2024autoregressive}
Sucheng Ren, Xianhang Li, Haoqin Tu, Feng Wang, Fangxun Shu, Lei Zhang, Jieru Mei, Linjie Yang, Peng Wang, Heng Wang, et~al.
\newblock Autoregressive pretraining with mamba in vision.
\newblock \emph{arXiv preprint arXiv:2406.07537}, 2024.

\bibitem[Tan and Le(2019)]{efficientnet}
Mingxing Tan and Quoc Le.
\newblock Efficientnet: Rethinking model scaling for convolutional neural networks.
\newblock In \emph{{ICML}}, 2019.

\bibitem[Touvron et~al.(2021)Touvron, Cord, Douze, Massa, Sablayrolles, and J{\'e}gou]{deit}
Hugo Touvron, Matthieu Cord, Matthijs Douze, Francisco Massa, Alexandre Sablayrolles, and Herv{\'e} J{\'e}gou.
\newblock Training data-efficient image transformers \& distillation through attention.
\newblock In \emph{{ICML}}, 2021.

\bibitem[Uy et~al.(2019)Uy, Pham, Hua, Nguyen, and Yeung]{ScanObjectNN19}
Mikaela~Angelina Uy, Quang-Hieu Pham, Binh-Son Hua, Thanh Nguyen, and Sai-Kit Yeung.
\newblock Revisiting point cloud classification: A new benchmark dataset and classification model on real-world data.
\newblock In \emph{IEEE/CVF Conf. Comput. Vis. Pattern Recog. (CVPR)}, pages 1588--1597, 2019.

\bibitem[Vaswani et~al.(2017)Vaswani, Shazeer, Parmar, Uszkoreit, Jones, Gomez, Kaiser, and Polosukhin]{AttentionIsAllYouNeed}
Ashish Vaswani, Noam Shazeer, Niki Parmar, Jakob Uszkoreit, Llion Jones, Aidan~N. Gomez, Lukasz Kaiser, and Illia Polosukhin.
\newblock Attention is all you need.
\newblock In \emph{Adv. Neural Inform. Process. Syst. (NeurIPS)}, pages 5998--6008, 2017.

\bibitem[Wang et~al.(2024{\natexlab{a}})Wang, Wang, Ren, Wei, Mei, Shao, Zhou, Yuille, and Xie]{wang2024mamba}
Feng Wang, Jiahao Wang, Sucheng Ren, Guoyizhe Wei, Jieru Mei, Wei Shao, Yuyin Zhou, Alan Yuille, and Cihang Xie.
\newblock Mamba-r: Vision mamba also needs registers.
\newblock \emph{arXiv preprint arXiv:2405.14858}, 2024{\natexlab{a}}.

\bibitem[Wang et~al.(2021{\natexlab{a}})Wang, Liu, Yue, Lasenby, and Kusner]{OcCo}
Hanchen Wang, Qi Liu, Xiangyu Yue, Joan Lasenby, and Matt~J Kusner.
\newblock Unsupervised point cloud pre-training via occlusion completion.
\newblock In \emph{Int. Conf. Comput. Vis. (ICCV)}, pages 9782--9792, 2021{\natexlab{a}}.

\bibitem[Wang et~al.(2021{\natexlab{b}})Wang, Liu, Yue, Lasenby, and Kusner]{wang2021unsupervised}
Hanchen Wang, Qi Liu, Xiangyu Yue, Joan Lasenby, and Matt~J Kusner.
\newblock Unsupervised point cloud pre-training via occlusion completion.
\newblock In \emph{Porc. of IEEE Intl. Conf. on Computer Vision}, 2021{\natexlab{b}}.

\bibitem[Wang et~al.(2024{\natexlab{b}})Wang, Song, Chen, Zhang, and Wang]{wang2024longllava}
Xidong Wang, Dingjie Song, Shunian Chen, Chen Zhang, and Benyou Wang.
\newblock Longllava: Scaling multi-modal llms to 1000 images efficiently via hybrid architecture.
\newblock \emph{arXiv preprint arXiv:2409.02889}, 2024{\natexlab{b}}.

\bibitem[Wang et~al.(2019)Wang, Sun, Liu, Sarma, Bronstein, and Solomon]{DGCNN}
Yue Wang, Yongbin Sun, Ziwei Liu, Sanjay~E. Sarma, Michael~M. Bronstein, and Justin~M. Solomon.
\newblock Dynamic graph {CNN} for learning on point clouds.
\newblock \emph{ACM Trans. Graph.}, 38\penalty0 (5):\penalty0 146:1--146:12, 2019.

\bibitem[Wang et~al.(2022)Wang, Yu, Rao, Zhou, and Lu]{P2P}
Ziyi Wang, Xumin Yu, Yongming Rao, Jie Zhou, and Jiwen Lu.
\newblock {P2P:} tuning pre-trained image models for point cloud analysis with point-to-pixel prompting.
\newblock In \emph{Adv. Neural Inform. Process. Syst. (NeurIPS)}, 2022.

\bibitem[Wu et~al.(2015)Wu, Song, Khosla, Yu, Zhang, Tang, and Xiao]{modelnet}
Zhirong Wu, Shuran Song, Aditya Khosla, Fisher Yu, Linguang Zhang, Xiaoou Tang, and Jianxiong Xiao.
\newblock 3d shapenets: A deep representation for volumetric shapes.
\newblock In \emph{Proceedings of the IEEE conference on computer vision and pattern recognition}, pages 1912--1920, 2015.

\bibitem[Xiao et~al.(2018)Xiao, Liu, Zhou, Jiang, and Sun]{xiao2018unified}
Tete Xiao, Yingcheng Liu, Bolei Zhou, Yuning Jiang, and Jian Sun.
\newblock Unified perceptual parsing for scene understanding.
\newblock In \emph{Proceedings of the European conference on computer vision (ECCV)}, pages 418--434, 2018.

\bibitem[Xie et~al.(2020)Xie, Gu, Guo, Qi, Guibas, and Litany]{PointContrast20}
Saining Xie, Jiatao Gu, Demi Guo, Charles~R. Qi, Leonidas~J. Guibas, and Or Litany.
\newblock Pointcontrast: Unsupervised pre-training for 3d point cloud understanding.
\newblock In \emph{Eur. Conf. Comput. Vis. (ECCV)}, pages 574--591. Springer, 2020.

\bibitem[Yi et~al.(2016)Yi, Kim, Ceylan, Shen, Yan, Su, Lu, Huang, Sheffer, and Guibas]{shapenetpart}
Li Yi, Vladimir~G Kim, Duygu Ceylan, I-Chao Shen, Mengyan Yan, Hao Su, Cewu Lu, Qixing Huang, Alla Sheffer, and Leonidas Guibas.
\newblock A scalable active framework for region annotation in 3d shape collections.
\newblock \emph{ACM Transactions on Graphics (ToG)}, 35\penalty0 (6):\penalty0 1--12, 2016.

\bibitem[Yu et~al.(2022{\natexlab{a}})Yu, Tang, Rao, Huang, Zhou, and Lu]{PointBERT}
Xumin Yu, Lulu Tang, Yongming Rao, Tiejun Huang, Jie Zhou, and Jiwen Lu.
\newblock Point-bert: Pre-training 3d point cloud transformers with masked point modeling.
\newblock In \emph{IEEE/CVF Conf. Comput. Vis. Pattern Recog. (CVPR)}, 2022{\natexlab{a}}.

\bibitem[Yu et~al.(2022{\natexlab{b}})Yu, Tang, Rao, Huang, Zhou, and Lu]{yu2022point}
Xumin Yu, Lulu Tang, Yongming Rao, Tiejun Huang, Jie Zhou, and Jiwen Lu.
\newblock Point-bert: Pre-training 3d point cloud transformers with masked point modeling.
\newblock In \emph{Proc. of IEEE Intl. Conf. on Computer Vision and Pattern Recognition}, 2022{\natexlab{b}}.

\bibitem[Zha et~al.(2023)Zha, Wang, Dai, Chen, Wang, and Xia]{zha2023instance}
Yaohua Zha, Jinpeng Wang, Tao Dai, Bin Chen, Zhi Wang, and Shu-Tao Xia.
\newblock Instance-aware dynamic prompt tuning for pre-trained point cloud models.
\newblock In \emph{Porc. of IEEE Intl. Conf. on Computer Vision}, 2023.

\bibitem[Zhang et~al.(2022)Zhang, Guo, Gao, Fang, Zhao, Wang, Qiao, and Li]{zhang2022point}
Renrui Zhang, Ziyu Guo, Peng Gao, Rongyao Fang, Bin Zhao, Dong Wang, Yu Qiao, and Hongsheng Li.
\newblock Point-m2ae: multi-scale masked autoencoders for hierarchical point cloud pre-training.
\newblock In \emph{Proc. of Advances in Neural Information Processing Systems}, 2022.

\bibitem[Zhang et~al.(2024)Zhang, Li, Yuan, Ji, and Yan]{PCM24}
Tao Zhang, Xiangtai Li, Haobo Yuan, Shunping Ji, and Shuicheng Yan.
\newblock Point could mamba: Point cloud learning via state space model.
\newblock \emph{arXiv preprint arXiv:2403.00762}, 2024.

\bibitem[Zhou et~al.(2017)Zhou, Zhao, Puig, Fidler, Barriuso, and Torralba]{zhou2017scene}
Bolei Zhou, Hang Zhao, Xavier Puig, Sanja Fidler, Adela Barriuso, and Antonio Torralba.
\newblock Scene parsing through ade20k dataset.
\newblock In \emph{Proceedings of the IEEE conference on computer vision and pattern recognition}, pages 633--641, 2017.

\bibitem[Zhu et~al.()Zhu, Liao, Zhang, Wang, Liu, and Wang]{zhuvision}
Lianghui Zhu, Bencheng Liao, Qian Zhang, Xinlong Wang, Wenyu Liu, and Xinggang Wang.
\newblock Vision mamba: Efficient visual representation learning with bidirectional state space model.
\newblock In \emph{Forty-first International Conference on Machine Learning}.

\bibitem[Zhu et~al.(2024)Zhu, Liao, Zhang, Wang, Liu, and Wang]{vim}
Lianghui Zhu, Bencheng Liao, Qian Zhang, Xinlong Wang, Wenyu Liu, and Xinggang Wang.
\newblock Vision mamba: Efficient visual representation learning with bidirectional state space model.
\newblock \emph{arXiv preprint arXiv:2401.09417}, 2024.

\end{thebibliography}
}

\clearpage
\setcounter{page}{1}
\maketitlesupplementary

\section*{Additional Experiments on 3D Tasks}
To verify that our method not only performs well on 2D tasks but is also effective for 3D tasks, we conducted experiments on additional 3D tasks. In addition to the ModelNet40\cite{modelnet} classification experiment, we also tested on the more challenging ScanObjectNN\cite{ScanObjectNN19} dataset. Unlike ModelNet40, which is a synthetic dataset, ScanObjectNN is a real-world object dataset and is commonly evaluated under three settings: OBJ\_BG, OBJ\_ONLY, and PB\_T50\_RS. Among these, the PB\_T50\_RS setting is the most challenging. Comparing the results of HybridNet and Mamba3D under the Supervised Learning Only setting reveals that HybridNet performs only slightly better than Mamba3D. However, both HybridNet and Mamba3D achieve significant performance improvements after MAP pretraining. This further validates that the MAP pretraining strategy is not only effective for hybrid frameworks but also enhances the pure Mamba framework. Comparing the results of Mamba3D under Point-BERT, Point-MAE, and MAP, it is evident that MAP demonstrates a significant performance advantage. This proves that even within the pure Mamba framework, MAP can surpass the performance of BERT-style and MAE-style pertaining.
We also conducted experiments on the few-shot learning task of ModelNet40 to validate the effectiveness of MAP. After MAP pretraining, both HybridNet and Mamba3D achieved significant performance improvements. On the more fine-grained task of ShapeNetPart\cite{shapenetpart} part segmentation, we also demonstrated that MAP can bring significant performance improvements to both hybrid frameworks and the pure Mamba framework.

\begin{table}[h]
    \renewcommand{\arraystretch}{1.1}
    \centering
    \resizebox{0.8\linewidth}{!}{
    \begin{tabular}{lcccc}
    \toprule[0.98pt]
    \multirow{2}{*}[-0.5ex]{Method}  & \multicolumn{2}{c}{5-way} & \multicolumn{2}{c}{10-way}\\
    \cmidrule(lr){2-3}\cmidrule(lr){4-5}  
    & 10-shot $\uparrow$ & 20-shot $\uparrow$ & 10-shot $\uparrow$ & 20-shot $\uparrow$ \\
    \midrule[0.6pt]
    \multicolumn{5}{c}{\textit{Supervised Learning Only} }\\
    \midrule[0.6pt]
    \bs DGCNN~\citep{DGCNN}   &31.6 {\small $\pm$ 2.8} &  40.8 {\small $\pm$ 4.6}&  19.9 {\small $\pm$ 2.1}& 16.9 {\small $\pm$ 1.5}\\
    \bv Transformer~\citep{AttentionIsAllYouNeed}   &87.8 {\small $\pm$ 5.2} & 93.3 {\small $\pm$ 4.3} &84.6 {\small $\pm$ 5.5} &89.4 {\small $\pm$ 6.3}\\
    \br{Mamba3D} \citep{han2024mamba3d} & {92.6 {\small $\pm$ 3.7}} & {96.9 {\small $\pm$ 2.4}} & {88.1 {\small $\pm$ 5.3}} & {93.1 {\small $\pm$ 3.6}} \\
    \rowcolor{cyan!10}
    \br{HybridNet}  & {92.8 {\small $\pm$ 3.2}} & {97.0 {\small $\pm$ 1.8}} & {88.4 {\small $\pm$ 4.3}} & {93.1 {\small $\pm$ 3.8}} \\
    \midrule[0.6pt]
    \multicolumn{5}{c}{\textit{with Self-supervised pretraining} }\\
    \midrule[0.6pt]
    \bs DGCNN+\textit{OcCo}\citep{OcCo}  &90.6 {\small $\pm$ 2.8} & 92.5 {\small $\pm$ 1.9} &82.9 {\small $\pm$ 1.3} &86.5 {\small $\pm$ 2.2}\\
    \bv OcCo~\citep{OcCo}  & 94.0 {\small $\pm$ 3.6}& 95.9 {\small $\pm$ 2.7 }&  89.4 {\small $\pm$ 5.1} & 92.4 {\small $\pm$4.6}\\
    \br PointMamba~\citep{PointMamba} & 95.0 {\small $\pm$ 2.3} & 97.3 {\small $\pm$ 1.8} &  91.4 {\small $\pm$ 4.4} & 92.8 {\small $\pm$ 4.0}\\
    \bv MaskPoint~\citep{MaskPoint}  & 95.0 {\small $\pm$ 3.7} & 97.2 {\small $\pm$ 1.7} & 91.4 {\small $\pm$ 4.0} & 93.4 {\small $\pm$ 3.5}\\
    \bv Point-BERT~\citep{PointBERT}  & 94.6 {\small $\pm$ 3.1} & 96.3 {\small $\pm$ 2.7} &  91.0 {\small $\pm$ 5.4} & 92.7 {\small $\pm$ 5.1}\\
    \bv Point-MAE~\citep{PointMAE} & 96.3 {\small $\pm$ 2.5}&97.8{\small $\pm$ 1.8} & {92.6 {\small $\pm$4.1}} & {95.0 {\small $\pm$ 3.0}}\\
    \br{Mamba3d}+\textit{P-B}~\citep{PointBERT} & {95.8 {\small $\pm$ 2.7}} & {97.9 {\small $\pm$ 1.4}} & {91.3 {\small $\pm$ 4.7}} & {94.5 {\small $\pm$ 3.3}} \\ 
    \br{Mamba3d}+\textit{P-M}~\citep{PointMAE} & {96.4 {\small $\pm$ 2.2}} & {98.2 {\small $\pm$1.2}} & {92.4 {\small $\pm$ 4.1}} & {95.2 {\small $\pm$ 2.9}} \\
    \rowcolor{cyan!10}
    \br{Mamba3d}+\textit{MAP} & {97.1 {\small $\pm$ 3.1}} & {98.7 {\small $\pm$1.3}} & {92.8 {\small $\pm$ 2.1}} & {95.8 {\small $\pm$ 3.1}} \\
    \rowcolor{cyan!10}
    \br{HybridNet}+\textit{MAP} & {97.3 {\small $\pm$2.8}} & {98.7 {\small $\pm$0.8}} & {93.0 {\small $\pm$ 3.6}} & {96.0 {\small $\pm$ 2.7}} \\
    \bottomrule[0.98pt]
    \end{tabular}
    }
    \vspace{-3mm}
    \caption{Few-shot classification on ModelNet40 dataset. The overall accuracy (\%) without voting is reported. \textit{P-B} and \textit{P-M} represent Point-BERT and Point-MAE strategy, respectively.}
    \label{tab:modelnet40}
\end{table}

\begin{table}[t]
\renewcommand{\arraystretch}{1.00}
   
    \label{tab:cls_main}
    \tiny
    \begin{center}
    \resizebox{0.98\linewidth}{!}{
        \begin{tabularx}{\linewidth}{>{\hsize=11\hsize}Xc@{\hspace{-0.2mm}}ccc@{\hspace{-0.1mm}}c@{\hspace{-0.1mm}}c}
        \toprule
        \multirow{2}{*}[-0.5ex]{Method} & \multirow{2}{*}[-0.5ex]{\texttt{PT}} & \multirow{2}{*}[-0.5ex]{\#P $\downarrow$} & \multirow{2}{*}[-0.5ex]{\#F $\downarrow$}  &  \multicolumn{3}{c}{ScanObjectNN} \\
        \cmidrule(lr){5-7} & & & & OBJ\_BG $\uparrow$ & OBJ\_ONLY $\uparrow$ & PB\_T50\_RS {$\uparrow$} \\
        \midrule[0.4pt]
        \multicolumn{7}{c}{\textit{Supervised Learning Only: Dedicated Architectures}}\\
        \midrule[0.4pt]
        PointNet\citep{PointNet} & $\times$ & 3.5 & 0.5 &  73.3 & 79.2 & 68.0 \\
        PointNet++\citep{PointNet++}  & $\times$ & 1.5 & 1.7   & 82.3 & 84.3 & 77.9 \\
        DGCNN\citep{DGCNN} & $\times$ & 1.8 & 2.4   & 82.8 & 86.2 & 78.1 \\
        PointCNN\citep{PointCNN} & $\times$ & 0.6 & - & 86.1 & 85.5 & 78.5 \\
        DRNet \citep{DRNet}& $\times$ & - & -  & - & -  & 80.3 \\
        SimpleView\citep{SimpleView} & $\times$ & - & -  & - & - & 80.5$\pm$0.3 \\
        GBNet\citep{GBNet} & $\times$ & 8.8 & -  & - & - & 81.0 \\
        PRA-Ne\citep{PRANet} & $\times$ & - & 2.3 & - & - & 81.0 \\
        MVTN\citep{MVTN} & $\times$ & 11.2 & 43.7 & {92.6} & {92.3} & 82.8 \\
        PointMLP\citep{PointMLP} & $\times$ & 12.6 & 31.4  & - & - & 85.4$\pm$0.3 \\
        PointNeXt\citep{PointNeXt} & $\times$ & 1.4 & 3.6  & - & - & 87.7$\pm$0.4 \\
        {P2P-HorNet}\citep{P2P} & $\checkmark$ & - & 34.6  & - & - & 89.3 \\
        DeLA\citep{DeLA} & $\times$ & 5.3 & 1.5 & - & - & {90.4} \\
        \midrule[0.4pt]
        \multicolumn{7}{c}{\textit{Supervised Learning Only: Transformer or Mamba-based Models}}\\
        \midrule[0.4pt]
        {Transformer} & $\times$ & 22.1 & 4.8 & 79.86 & 80.55 & 77.24 \\
        PCT\citep{PCT} & $\times$ & 2.9 & 2.3  & - & - & - \\
        {PointMamba}\citep{liang2024pointmamba} & $\times$ & 12.3 & 3.6 & 88.30 & 87.78 & 82.48 \\
        PCM\citep{PCM24} & $\times$ & 34.2 & 45.0 & - & - & 88.10$\pm$0.3 \\
        SPoTr\citep{SPoTr} & $\times$ & 1.7 & 10.8  & - & - & 88.60 \\
        PointConT\citep{PointConT} & $\times$ & - & - & - & - & 90.30 \\
        {Mamba3d w/o vot.}\citep{han2024mamba3d} & $\times$ & 16.9 & 3.9  & {92.94} & {92.08} & {91.81}  \\
        {Mamba3d w/ vot.}\citep{han2024mamba3d} & $\times$ & 16.9 & 3.9  & {94.49}  & {92.43}  & {92.64}  \\\rowcolor{cyan!10}
        {HybridNet w/o vot.} & $\times$ & 19.3 & 4.4  & {92.81} & {92.28} & {91.97}  \\\rowcolor{cyan!10}
        {HybridNet w/ vot.} & $\times$ & 19.3 & 4.4  & {94.50}  & {92.58}  & {92.66}  \\
        \midrule[0.4pt]
        \multicolumn{7}{c}{\textit{With Self-supervised pretraining} }\\
        \midrule[0.4pt]
        Transformer & \textit{OcCo} & 22.1 & 4.8 & 84.85 & 85.54 & 78.79 \\
        Point-BERT & \textit{IDPT} & 22.1+1.7$^\dagger$ & 4.8 & 88.12 & 88.30 & 83.69 \\
        MaskPoint & \textit{MaskPoint} & 22.1 & 4.8 & 89.30 & 88.10 & 84.30 \\
        PointMamba & \textit{Point-MAE} & 12.3 & 3.6 & 90.71 & 88.47 & 84.87 \\
        Point-MAE & \textit{IDPT} & 22.1+1.7$^\dagger$ & 4.8  & 91.22 & 90.02 & 84.94 \\
        Point-M2AE & \textit{Point-M2AE} & 15.3 & 3.6  & 91.22 & 88.81 & 86.43 \\
        {Mamba3d w/o vot.} & \textit{Point-BERT} & 16.9 & 3.9  & {92.25} & {91.05} & {90.11} \\
        Point-MAE & \textit{Point-MAE} & 22.1 & 4.8  & 90.02 & 88.29 & 85.18 \\
        {Mamba3d w/o vot.} & \textit{Point-MAE}  & 16.9 & 3.9 & {93.12} & {92.08} & {92.05} \\
        {Mamba3d w/ vot.} & \textit{Point-MAE}  & 16.9 & 3.9 & {95.18} & {94.15} & {93.05} \\\rowcolor{cyan!10}
        {Mamba3d w/o vot.} & \textit{MAP}  & 16.9 & 3.9 & {93.62} & {92.75} & {92.65} \\\rowcolor{cyan!10}
        {Mamba3d w/ vot.} & \textit{MAP}  & 16.9 & 3.9 & {95.64} & {94.87} & {93.76} \\\rowcolor{cyan!10}
        {HybridNet w/o vot.} & \textit{MAP}  & 19.3 & 4.4 & {93.88} & {93.03} & {92.95} \\\rowcolor{cyan!10}
        {HybridNet w/ vot.} & \textit{MAP}  & 19.3 & 4.4 & {95.84} & {94.97} & {93.87} \\
        \bottomrule[0.7pt]
        \end{tabularx}
    }
    \vspace{-3mm}
     \caption{Results on 3D classification tasks. Our results are highlighted in \colorbox{cyan!10}{blue}. PT: pre-training strategy. }
     \vspace{-5mm}
    \end{center}
\end{table}

\begin{table}[t]
    \renewcommand{\arraystretch}{1.1}
    \vspace{-5pt}
    \centering
    \resizebox{0.8\linewidth}{!}{
    \begin{tabular}{lcccc}
    \toprule[0.8pt]
    Method & mIoU$_C$ (\%) $\uparrow$ & mIoU$_I$ (\%) $\uparrow$ & \#P $\downarrow$ & \#F $\downarrow$ \\
    \midrule[0.4pt]
    \multicolumn{5}{c}{\textit{Supervised Learning Only} }\\
    \midrule[0.4pt]
    \bs PointNet~\citep{PointNet}  & 80.4 & 83.7 & 3.6 & 4.9 \\
    \bs PointNet++~\citep{PointNet++} & 81.9 & 85.1 & {1.0} & {4.9} \\
    \bs DGCNN~\citep{DGCNN} & 82.3 & 85.2 & 1.3 & 12.4 \\
    \bv Transformer~\citep{AttentionIsAllYouNeed} & 83.4 & 85.1 & 27.1 & 15.5 \\
    \br{Mamba3D}\citep{han2024mamba3d} & {83.7} & {85.7} & 23.0 & 11.8 \\\rowcolor{cyan!10}
    \br{HybridNet} & {83.5} & {85.6} & 25.1 & 12.9 \\
    \midrule[0.4pt]
    \multicolumn{5}{c}{\textit{with Self-supervised pretraining} }\\
    \midrule[0.4pt]
    \bv OcCo~\citep{OcCo} & 83.4 & 84.7 & 27.1 & - \\
    \bs PointContrast~\citep{PointContrast20} & - & 85.1 & 37.9 & - \\
    \bs CrossPoint~\citep{CrossPoint22} & - & 85.5 & - & - \\
    \midrule[0.4pt]
    \bv Point-MAE~\citep{PointMAE} & {84.2} & {86.1} & 27.1 & 15.5 \\
    \br PointMamba~\citep{PointMamba} & {84.4} & 86.0 & {17.4} & 14.3\\
    \bv Point-BERT~\citep{PointBERT} & 84.1 & 85.6 & 27.1 & 10.6 \\
    \br{Mamba3d}+\textit{P-B} ~\citep{PointBERT} & {84.1} & {85.7} & {21.9} & {9.5} \\
    \br{Mamba3d}+\textit{P-M}~\citep{PointMAE} & {84.3} & {85.8} & 23.0 & 11.8 \\\rowcolor{cyan!10}
    \br{Mamba3d}+\textit{MAP} & {84.5} & {86.0} & 23.0 & 11.8 \\\rowcolor{cyan!10}
    \br{HybridNet}+\textit{MAP} & {84.7} & {86.3} & 25.1 & 12.9 \\
\bottomrule[0.8pt]
    \end{tabular}
    }
    \vspace{-3mm}
    \caption{Part segmentation on ShapeNetPart dataset. Our results are highlighted in \colorbox{cyan!10}{blue}. The class mIoU (mIoU$_C$) and the instance mIoU (mIoU$_I$) are reported, with model parameters \#P (M) and FLOPs \#F (G).}
\end{table}

\end{document}